\pdfoutput=1

\documentclass[pdflatex,sn-basic]{sn-jnl}

\jyear{2022}%

\theoremstyle{thmstyleone}%
\theoremstyle{thmstyletwo}%

\theoremstyle{thmstylethree}%

\usepackage{graphicx}
\usepackage{amsmath}
\usepackage{amssymb}
\usepackage{caption}
\usepackage{float}
\usepackage{multirow}
\usepackage[normalem]{ulem}
\useunder{\uline}{\ul}{}
\usepackage{booktabs}
\usepackage{subfigure}
\usepackage{pdfpages}
\usepackage{comment}

\begin{document}

\title[Learning in the home cage using dual-stream spatiotemporal networks]{Dual-stream spatiotemporal networks with feature sharing for monitoring animals in the home cage}


\author*[1]{\fnm{\hspace{3em}Ezechukwu I.} \sur{Nwokedi}}\email{enwokedi@lincoln.ac.uk}

\author[2]{\fnm{Rasneer S.} \sur{Bains }}\email{r.bains@har.mrc.ac.uk}

\author[1]{\\ \fnm{Luc} \sur{Bidaut}}\email{lbidaut@lincoln.ac.uk}

\author[1]{\fnm{Xujiong} \sur{Ye}}\email{xye@lincoln.ac.uk}

\author[2]{\fnm{Sara} \sur{Wells}}\email{s.wells@har.mrc.ac.uk}

\author[1]{\fnm{James M.} \sur{Brown}}\email{jamesbrown@lincoln.ac.uk}

\affil*[1]{\orgdiv{School of Computer Science}, \orgname{University of Lincoln}, \orgaddress{\street{Brayford Way}, \city{Lincolnshire},   \country{UK}}}

\affil[2]{\orgdiv{Mary Lyon Centre}, \orgname{MRC Harwell Instituite}, \orgaddress{\street{Harwell}, \city{Oxfordshire},   \country{UK}}}


\abstract{This paper presents a spatiotemporal deep learning approach for mouse behavioural classification in the home-cage. Using a series of dual-stream architectures with assorted modifications to increase performance, we introduce a novel \emph{feature sharing} approach that jointly processes the streams at regular intervals throughout the network. 
Using an annotated, publicly available dataset of a singly-housed mice, we achieve prediction accuracy of 86.47\%\ using an ensemble of a Inception-based network and an attention-based network, both of which utilize this\emph{feature sharing}. We also demonstrate through ablation studies that for all models, the \emph{feature sharing} architectures consistently perform better than conventional ones having separate streams. The best performing models were further evaluated on other activity datasets, both mouse and human. Future work will investigate the effectiveness of \emph{feature sharing} to behavioural classification in the unsupervised anomaly detection domain.}






\maketitle

\section{Introduction}\label{sec1}

Over many decades, the ethical implications of animals in research has undergone considerable discussion and scrutiny \citep{akhtar2015flaws}. 
A major landmark geared at regulating and improving the use of animals in research is the 3R’s - Replacement, Refinement and Reduction - and the National Centre for 3R’s that spearheads its movement in the UK. As of 2020, statistics show that around 2.88 million living animals were used for various research procedures in the UK, with 92\%\ of these made up of rodents and fish \citep{NC3Rs2022animals}. Due to their genetic and anatomical similarities with humans, rodents (such as mice and rats) are some of the most utilized animals in biomedical research.

In support of the 3Rs mission, technology has been increasingly used to quantify different aspects of research involving animals. Behavioural phenotyping is particularly important as it is also the primary means of determining welfare concerns that may arise during an experiment. However, the manual detection of these behaviours is expensive, laborious, time consuming and overall, was not easily reproducible \citep{karl2003behavioral,jhuang2010automated}. The development of home-cage monitoring (HCM) systems is a major technological solution which has helped to solve many of these issues \citep{voikar2020three}. HCM systems facilitate non-intrusive observation of mice and may provide a range of functions such as behavioural annotation and ethogramming, depth sensing and tracking, activity summarising of circadian rhythm, and pose estimation. These HCM systems include the Techniplast Digital Ventilated Cage (DVC) \citep{iannello2019non}, System for Continuous Observation of Rodents in Home-cage Environment (SCORHE) \citep{salem2015scorhe} and IntelliCage \citep{kiryk2020intellicage}, to name a few. However, there are few commercially available solutions to the problem of detecting behaviours from video footage alone. Moreover, many of the solutions that do exist are strongly coupled to commercial hardware, rather than video footage in general.

\begin{figure}
    \centering
    \captionsetup{justification=centering}
     \subfigure[Conventional template of dual stream architectures]{\label{normieTemplate}\includegraphics[width=\textwidth,keepaspectratio, page=1]{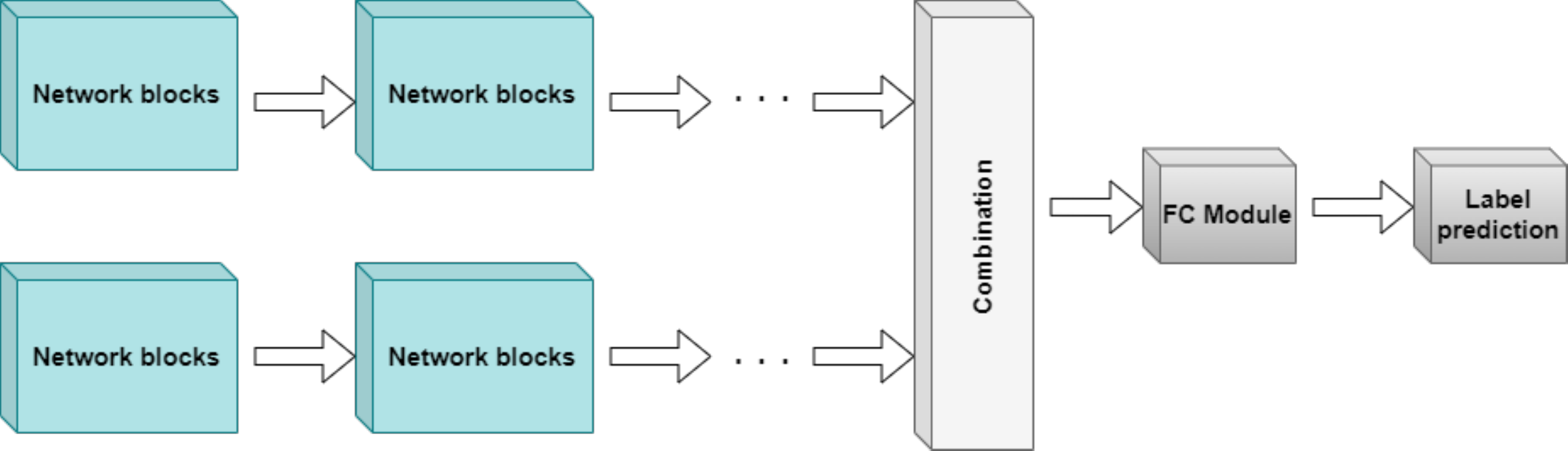}}

     \subfigure[Our proposed dual stream template]{\label{revisedTemplate}\includegraphics[width=\textwidth, page=2]{figure_topmodules.pdf}}
    \caption{Conventional standalone vs feature-sharing dual networks}
    \label{fig:topmodules}
\end{figure}

In this paper, dual-stream deep learning architectures are proposed for behavioural classification of mice in the home-cage. The models in question were developed for entirely supervised learning, whereby spatiotemporal (ST) blocks of video data are mapped to one of several behaviour categories. The dataset utilised is publicly available and contains videos of a singly-housed mice \citep{jhuang2010automated}. Their results however are not directly comparable to ours as the models here were trained and evaluated differently. More on this is explained in \ref{sec3_data}. One of the novel aspects of these models is shared layers between the streams of the networks.  Here, instead of fusing individual streams at the end \citep{simonyan2014two}, we propose to combine features at regular interval throughout the architecture. We hypothesize that accurate representations are better enforced when both streams are privy to information from each other (figure \ref{fig:topmodules}). Some instances of shared features has been seen in unets \citep{han2017automatic} and its many derivative networks, and some other specialized multi-stream architectures \citep{zhang2020local,hou2021local} however not in the same manner as proposed here for multi-stream networks.

\section{Related Works}\label{sec2}


\subsection{Video classification} 
In the work by the orginal creators of the dataset used in this paper \citep{jhuang2010automated}, the authors presented a support vector machine (SVM) classifier coupled with a Hidden Markov Model (HMM) for sequence tagging. Main features extracted from videos were the mouse position and motion statistics, which were consequently mapped to the mouse behaviour(s). The model training was repeated n=12 times using a \emph{leave-one-out} methodology, achieving a classification accuracy of 77.3\%\ across all 8 classes as opposed to the 71.6\%\ accuracy by human annotators.

Since then, deep learning has emerged as the state-of-the-art for classification of video data. Though 2D models thrived in most applications, the need for better contextual understanding has become increasingly apparent, and the application of 3D convolutions has enabled this. Better yet, the use of multiple streams allows for better encoding of video or clip sequences. It is often the case that one of the model streams operates on an image or image sequence (within time frame $t_0$ to $t_n$) while the next stream operates on the optical flow (computed for $t_1$ to $t_{n+1}$) \citep{carreira2017quo, simonyan2014two}. Some other multi-stream variations operate on two image streams of different point of views, resolutions \citep{wei2020two} or zooms depending on the goal of classification.

Work by \citep{carreira2017quo} presented a new network called the inflated 3D (I3D) Inception model. The I3D modules differed from the classic Inception module due to the addition of 3D convolutions and ‘inflated’ filters that allowed wider receptive field necessary to better learn spatiotemporal data. This dual stream I3D architecture (pretrained on ImageNet) was utilized by \citep{nguyen2019applying} to classify home cage mouse behaviours. Its evaluation was carried out on the same MIT dataset \citep{jhuang2010automated} and used a \emph{leave-one-out} method, therefore averaging test results across the twelve videos in the main dataset. They achieved an average accuracy of over 90\%\ on testing with various stream weights.

In another paper by \citep{hou2021local}, the effect of shared features at higher levels of multi-stream networks was demonstrated. The authors termed this operation \emph{feature fusion}. The architecture comprised of a framewise spatial transformer-based stream and a clipwise temporal stream. The stream features were combined at two successive final pooling layers. They achieved excellent results with accuracies of 95.3\% on the UCF101 \citep{soomro2012ucf101} and 72.9\% on the HMDB51 \citep{kuehne2011hmdb} datasets.  Note however that the \emph{feature sharing} proposed in our work is implemented at multiple points throughout the dual-stream architectures and leverages on the clipwise nature of both streams. More on \emph{feature sharing} is explained in section \ref{sec3Arch}.

\subsection{Spatiotemporal Learning}
Though utilized in a different machine learning algorithm, \citep{dollar2005behavior} proved that spatiotemporal cuboids of data formed better descriptors in both human activity recognition and mice behaviour classifications task. These spatiotemporal features were implied to be better in video classification due to the presence of more information which better captures event contexts, especially those that can be easily confused at instance classification. In rodent phenotyping, a lot of emphasis is placed on themes such as behavioural sequences, periodicity, repetitiveness, or patterns of certain exhibited behaviours \citep{kyzar2011towards}. Depending on the researchers’ goal, these become increasingly important, else subtle details are missed. A good example of this is self-grooming behaviour which can be observed as mice transition from their idle periods to high activity \citep{kalueff2005mouse, kyzar2011towards}. However, when in excess, this behaviour is also commonly associated with mice models of both autism spectrum disorders and compulsive disorders \citep{liu2021dissection}. This further attests to the importance of temporal content capturing temporal representations in machine learning models.

One of the key architectures to spatiotemporal learning is the I3D \citep{carreira2017quo} mentioned earlier. In its basic form, I3D was built by changing the 2D convolutional layers in the Inception v1 \citep{szegedy2015going} model to 3D convolutions while still leveraging on the efficient structure of the Inception blocks.  Unlike other 3D convolutional architectures, I3D are deep yet lightweight. In addition, Inception networks have well proven themselves in image classification, therefore it's expansion to learning temporal content was simply expanding its capabilities. Owing to its advantages, the same concept has been applied in some architectures proposed for this paper.


\subsubsection{Attention mechanisms} An attention module is characterized by the following elements: query Q, key K and value V. It attempts to map these to the output and scales the output using the dimension of the keys ${d_k}$ (seen in the dot-product version). Multi-head attention (MHA) combines multiple attention instances with trainable parameters W and is often utilized to ensure efficient learning of vector sequences  \citep{vaswani2017attention}. The general expressions for the attention function and multi-head attention are given below:

\begin{equation}Attention(Q,K,V)=softmax((QK^T)/\sqrt{(d_k )})V \end{equation} \
\begin{equation}MHA(Q,K,V)=Concatenate(head_1,head_2,…,head_i) W^o\end{equation} \

where $head_i=Attention(QW_i^Q,KW_i^K,VW_i^V)$.
\\

Transformers are a derivative architecture of MHA initially applied to natural language understanding \citep{vaswani2017attention} but have also been found to be effective in computer vision. The Vision Transformers (ViT) is one of these which repurposed transformers to image tasks \citep{dosovitskiy2020image}. Further variations of ViTs designed for spatiotemporal learning of videos have achieved state-of-the-art (SOTA) results in activity recognition \citep{bertasius2021space, arnab2021vivit}. The work by \citep{bertasius2021space} also proved that multi-head attention captures vital temporal dependencies by focusing on displaced or moving objects within a sequence. Furthermore, its application was proven to be effective in capturing global features in a multi-stream architecture for video classification \citep{li2020novel}.

\subsubsection{Long-Short Term Memory (LSTM)} LSTMs are architectures which learn to store information using memory cells and gates. The memory cell was designed to achieve constant error flow and used multiplicative input and output gates that protect data from perturbation \citep{hochreiter1997long}. Further improvements after this saw better defined gate operations which improved the memory retention of the architecture. 



Building upon this, Bidirectional LSTMs (BiLSTM) allow for computation of memory both ways and have been proven to achieve good results in vision tasks. BiLSTM are composed of two LSTMs that store relevant dependencies from both forward (i.e. past to present) and backwards (i.e. future to present) state directions \citep{gharagozloo2021machine}. In conjunction with other ML architectures, bidirectional LSTMs have been found to outperform the unidirectional LSTM in several natural language understanding \citep{graves2005bidirectional, suzuki2018convolutional} and image classification \citep{hua2019recurrently} tasks. In the paper by \citep{gharagozloo2021machine}, BiLSTM was used with 1-dimensional convolutions to classify the circadian rhythm of wild-type mice into day or night states. This was trained after the dimensionality reduction of a five-minute clip which was further subdivided into three-second frame windows. It was found to outperform the other ML algorithms explored, with area-under-the-curve (AUC) of 0.97. In short, BiLSTMs are capable of efficiently detecting and learning patterns that define the behaviours mapped.

\section{Methodology}\label{sec3}

\subsection{Data}\label{sec3_data}
The MIT mice dataset is subdivided into a main dataset and clipped database \citep{jhuang2010automated}. In this work, we utilize all twelve videos from the main dataset for training and validation while the clipped database, composed of unambiguous behaviours, is used to test the models. Unlike the \emph{leave-one-out} methodology by the original authors, we surmise that this approach helps to better examine the generalisation performance of our models. The optical flows generated from these was computed using dense optical flow method \citep{farneback2003two}. Both training and test frames were resized to $128 \times 128$, and further reduced to $128 \times 96$ by uniformly cropping redundant parts of each frame that lie along the vertical axis. The data was also temporally downsampled using five-frame intervals (that is, every second was represented by one-sixth of a second). The temporal length used for each \emph{T=8} frames. Thus, each spatiotemporal cuboid represents approximately 1.33 seconds of the original video. Towards the end of the videos/clips, any frames which could not fit these specifications were discarded. The final input data is in the form $\mathit{N} \times \mathit{T} \times \mathit{W} \times \mathit{H} \times \mathit{C}$ which represent the number of clips, temporal length, spatial width, spatial height and number of channels, respectively. The \emph{N} values for the final training, validation and testing sets are 23,444, 4,195 and 5,171 respectively.

\begin{figure}[h]
    \centering
    \captionsetup{justification=centering}
    \includegraphics{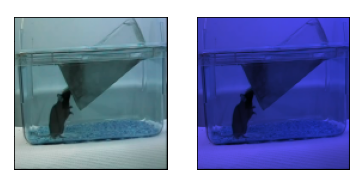}
    \caption{sample frame before and after ‘\emph{nightification}’}
    \label{fig:mice}
\end{figure}

In the task of classification, there are often prediction discrepancies associated with various inconsistencies. One of these is class imbalance (see appendix \ref{secA3} for datasets' distribution of frames to classes). Here, the severe imbalance in class distribution was alleviated using weights \citep{king2001logistic} which forced the model to percieve the number of samples in each class as having the same value. Hence, the classes which suffered from low sample sizes, such as \emph{drinking}, were assigned higher weights and the reverse for labels with large sample sizes like \emph{micromovement}. 

Another such inconsistency was the varying lengths of day and night videos. In this particular dataset, there are only two videos recorded in night-time (using infrared cameras) while all the rest (including the clipped dataset) were recorded in the day. In some deep learning applications, conversion to grayscale has proven effective but this method was found degraded the performance of the models. As such, all day videos contained within the dataset were ‘\emph{nightified}’ (i.e., changed into night-time). This was achieved by first calculating the averaged R, G, and B channel values from the two night videos. These were then used to weight the [0-1] normalised data from the day videos and finally expanded back to [0-255] range. The results gave a close approximation of what the videos would look like if recorded in the night, and thus lessened bias in the models caused by the day-night imbalance (figure \ref{fig:mice}). No further augmentations were performed on the dataset. More data samples, used in both RGB and flow streams, are available in appendix \ref{secA4}.

\subsection{Architectures} \label{sec3Arch}

\begin{figure}[H]
\centering
 \subfigure[Primary module I]{\label{pmod1}\includegraphics[scale=0.22, page=1,trim={0.5cm .5cm .5cm 2cm},clip,]{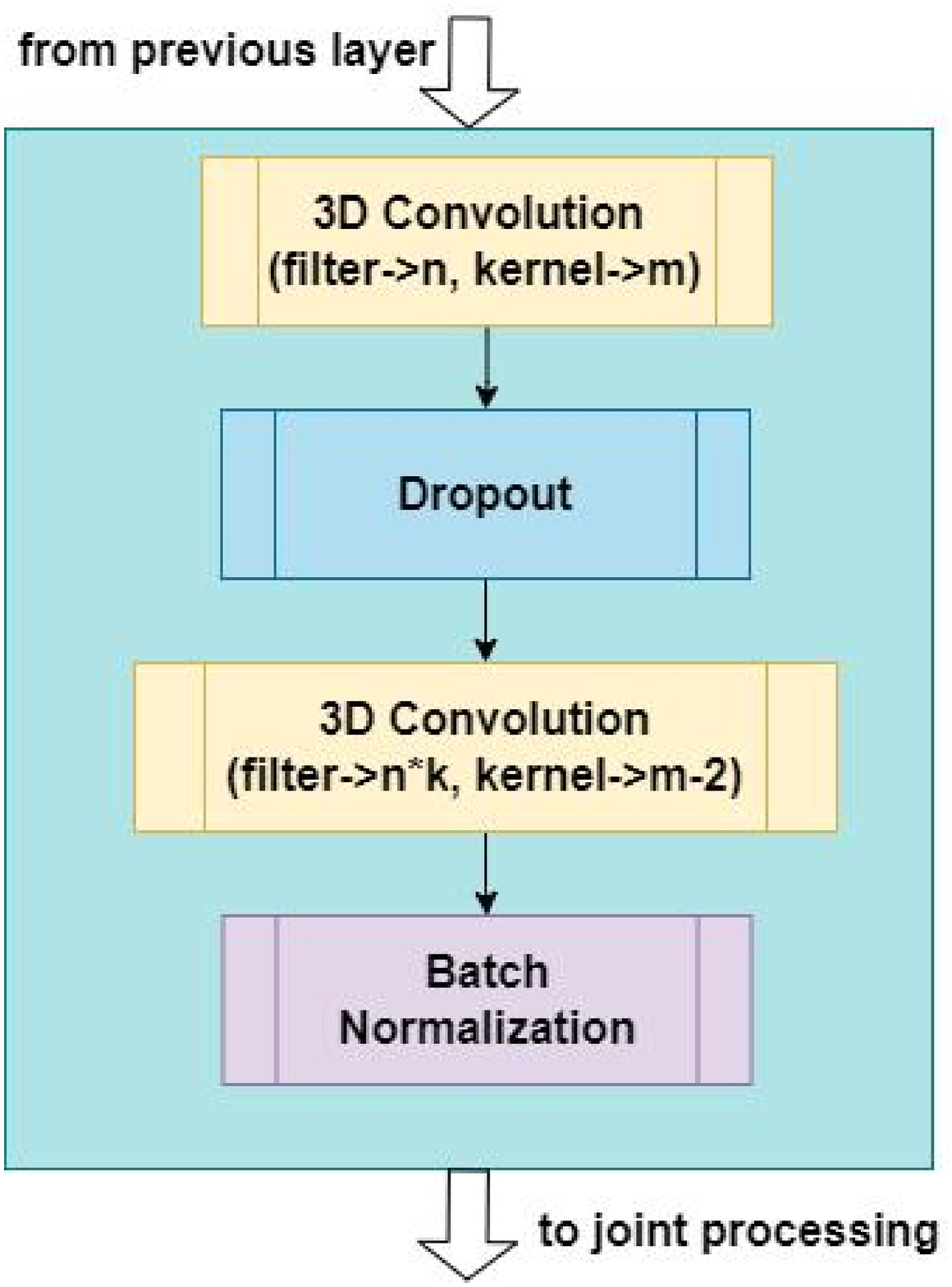}}
\subfigure[Joint processing module]{\label{inceptv3D}\includegraphics[scale=0.28, page=3]{figure_submodules-converted.pdf}}
\vspace{-.5cm}
\subfigure[Primary module II]{\label{pmod2}\includegraphics[width=0.63\textwidth, page=2,]{figure_submodules-converted.pdf}}
\subfigure[Simple joint processing]{\label{simp}\includegraphics[scale=0.15,keepaspectratio, page=4,trim={0.5cm 0cm .5cm .5cm},clip,]{figure_submodules-converted.pdf}}
\caption{Network modules used in dual-stream architectures.}
\label{fig:submodules}
\end{figure}

All of the models presented are multi-stream and, in this application, use raw video and optical flow streams. The building blocks utilized in the networks are depicted in figure \ref{fig:submodules}. As earlier stated in the introduction, one of the vital aspects of the models presented here is the \emph{feature sharing} between the dual-streams of the network. \emph{Feature sharing} entails combination and/or joint processing of the stream outputs after operation by the primary modules. This combination is achieved either via addition or concatenation, followed by further processing on the joint streams which are then projected back to the individual streams. These operations take place at regular intervals throughout the architectures.  We hypothesize that this procedure reinforces learnt features better than operating on the streams individually. The various implementations of this are further discussed under each architecture, and in table \ref{summaryFSMods}. The overview of each architecture is also depicted in appendix \ref{secA5}.

The variables \emph{k}, \emph{n} and \emph{m} used in figure \ref{fig:submodules} represent the multipliers, filter sizes and kernel sizes used at different levels in the architectures.

The blocks in subfigures \ref{pmod1} and \ref{pmod2} represent the primary processing modules used in both the RGB image and optical flow streams, while the blocks in \ref{inceptv3D} and \ref{simp} are the joint processing modules. The blocks in \ref{pmod2} and \ref{inceptv3D} depict 3D formats of modules originally found in Inception v3 and Inception v1 architectures respectively \citep{szegedy2015going, szegedy2016rethinking}. In particular, \ref{inceptv3D} was adapted here to boost the performance of the architectures utilizing module \ref{pmod1} via further processing at the junctions where the streams meet. Block \ref{simp} is a custom joint processing module utilized only in the baseline network.

\subsubsection{Baseline network (CNN)} This simple architecture consists of blocks with 3D convolutional layers, dropout (with uniform rates of 20\%),\ and batch normalization (see figure \ref{pmod1}). The kernel sizes here were made uniform for each block (i.e. kernels of size $m$ rather than $m-2$ as depicted in figure \ref{pmod1}). After operation by similar blocks, the results from both streams are summed up and further operated on by dense and dropout layers (figure \ref{simp}) before splitting again into the individual streams. 

\subsubsection{CNN + Inception v3\_D + Attention (CIv3D\_MHA)} This builds on the baseline architecture, adding the self-attention mechanisms to both streams after the last primary blocks. The kernel size for 3D convolutions were made to undulate (as shown in figure \ref{pmod1}) and inversely mirrored between parallel stream blocks. In addition, the simple processing block is replaced by the InceptionD module \citep{szegedy2016rethinking} (figure \ref{inceptv3D}) throughout the architecture. The self-attention block used here is similar to vision transformers \citep{dosovitskiy2020image} however it uses batch normalization, and the patch tokens are replaced by the end features of the streams before summation and processing by the last InceptionD block. 

\subsubsection{CNN + Inception v3\_D + BiLSTM (CIv3D\_BiLSTM)} This uses the same improvisations made in CIv3D\_MHA but removes the primary modules' dropout layers. The bidirectional LSTMs are used in place of the traditional flattening that precedes fully-connected layers. The input to this is the summed output of both streams’ final subsection, reshaped from four to two dimensions to allow loading into the LSTMs. 

\subsubsection{Purely Inception-based networks}\label{srs_crs}
There are two architectures completely built-up using the 3D Inception v1 block (see subfigure \ref{pmod2}). This block was revised for spatiotemporal operation from the dimensionality reduction module in the classic Inception v1 architecture \citep{szegedy2015going} but is without the singular $1 \times 1$ convolution branch in the original (figure \ref{pmod2}). The first architecture can be best described as multi-stream. At the bottleneck between successive subregions of the network, feature learning is reinforced by repeatedly combining strided computes of the original optical flow sequence with the previous features extracted from the RGB stream. Hence the network was termed Singly Reinforced Stream (SRS) network. It also adds an intricate detail of removing the first and last two frames of the optical flow stream (along with some surrounding dimensions) after the first block operation on both streams. This cropping operation is carried out only once and done under the assumption that the temporal sequence is better represented by the centre portions of the mid-four frames. This train of thought is quite similar to the fovea stream in \citep{karpathy2014large} but takes it further by removing frames at the extremities.

Unlike the SRS network, the second architecture was developed to encourage cross-pollination between streams; this implies that just as the optical stream enforces representation learning in the image stream, the image stream is also used to enforce learning in the optical stream, and they alternate in this manner. This is done by independently concatenating the past features from each streams’ block with the jointly-processed input fed into consequent blocks. This operation however led to a considerable increase in computation (see parameter count in Table \ref{tab}). This network was named Cross Reinforced Streams (CRS) network. 

\subsubsection{Other networks} To investigate the effectiveness of the shared layers between streams, experiments were conducted on versions of the above models without the unique \emph{feature sharing} modules. The hallmark algorithms used in each architecture were left in-situ while the areas of joint processing are replicated in both streams, all before the common fully-connected layers. 

\begin{table}[h]
\fontsize{17}{20}\selectfont
\centering
\caption{Full summary of feature sharing models}
\label{summaryFSMods}
\resizebox{\textwidth}{!}{%
\begin{tabular}{|lccccc|}
\hline
\multicolumn{1}{|l|}{\textbf{Models}} &
  \multicolumn{1}{c|}{Baseline} &
  \multicolumn{1}{c|}{CIv3D\_BiLSTM} &
  \multicolumn{1}{c|}{CIv3D\_MHA} &
  \multicolumn{1}{c|}{SRS} &
  CRS \\ \hline
\multicolumn{1}{|l|}{\textbf{Primary filters (n)\footnotemark}}
&
  \multicolumn{1}{l|}{16, 32, 64, 128} &
  \multicolumn{1}{c|}{16, 32, 64, 128} &
  \multicolumn{1}{c|}{16, 32, 64, 128} &
  \multicolumn{1}{c|}{\begin{tabular}[c]{@{}c@{}}8, 16, 32, \\ 64, 64,128\end{tabular}} &
  \begin{tabular}[c]{@{}c@{}}12, 24, 48, \\ 96, 192\end{tabular} \\ \hline
\multicolumn{1}{|l|}{\textbf{\begin{tabular}[c]{@{}l@{}}Intra-block \\ filter multipliers (k)\footnotemark[\value{footnote}] \end{tabular}}} &
  \multicolumn{1}{c|}{1.5} &
  \multicolumn{1}{c|}{1.5} &
  \multicolumn{1}{c|}{1.5} &
  \multicolumn{1}{c|}{\begin{tabular}[c]{@{}c@{}}1.5, 1.5, 1.5, \\ 1.5, 1.5, 1.0\end{tabular}} &
  1.0 \\ \hline
\multicolumn{1}{|l|}{\textbf{Kernels (m)}} &
  \multicolumn{1}{c|}{5, 3, 5, 5} &
  \multicolumn{1}{c|}{7, 5, 5, 3} &
  \multicolumn{1}{c|}{7, 5, 5, 3} &
  \multicolumn{1}{c|}{\begin{tabular}[c]{@{}c@{}}refer to \\ subfigure \ref{pmod2}\end{tabular}} &
  \multicolumn{1}{l|}{\begin{tabular}[c]{@{}c@{}}refer to \\ subfigure \ref{pmod2}\end{tabular}} \\ \hline
\multicolumn{1}{|l|}{\textbf{\begin{tabular}[c]{@{}l@{}}Stream combination \\ via?\end{tabular}}} &
  \multicolumn{1}{c|}{Addition} &
  \multicolumn{1}{c|}{Addition} &
  \multicolumn{1}{c|}{Addition} &
  \multicolumn{1}{c|}{Concatenation} &
  Concatenation \\ \hline
\multicolumn{1}{|l|}{\textbf{Processing at joints?}} &
  \multicolumn{1}{c|}{\begin{tabular}[c]{@{}c@{}}Yes (refer to \\ subfigure \ref{simp})\end{tabular}} &
  \multicolumn{1}{c|}{\begin{tabular}[c]{@{}c@{}}Yes (refer to \\ subfigure \ref{inceptv3D})\end{tabular}} &
  \multicolumn{1}{c|}{\begin{tabular}[c]{@{}c@{}}Yes (refer to \\ subfigure \ref{inceptv3D})\end{tabular}} &
  \multicolumn{1}{c|}{No} &
  No \\ \hline
\multicolumn{1}{|l|}{\textbf{\begin{tabular}[c]{@{}l@{}}Further processing \\ before FC?\end{tabular}}} &
  \multicolumn{1}{c|}{No} &
  \multicolumn{1}{c|}{No} &
  \multicolumn{1}{c|}{No} &
  \multicolumn{1}{c|}{\begin{tabular}[c]{@{}l@{}}Single Inception v1 \\ block(subfigure \ref{pmod2})\end{tabular}} &
  No \\ \hline
\multicolumn{1}{|l|}{\textbf{\begin{tabular}[c]{@{}l@{}}Activation function(s) \\ before FC?\end{tabular}}} &
  \multicolumn{1}{c|}{Leaky ReLU} &
  \multicolumn{1}{c|}{Leaky ReLU} &
  \multicolumn{1}{c|}{Leaky ReLU} &
  \multicolumn{1}{c|}{ReLU} &
  ReLU \\ \hline
\multicolumn{1}{|l|}{\textbf{\begin{tabular}[c]{@{}l@{}}Activation at \\ last dense layer?\end{tabular}}} &
  \multicolumn{1}{c|}{Softmax} &
  \multicolumn{1}{c|}{Softmax} &
  \multicolumn{1}{c|}{Softmax} &
  \multicolumn{1}{c|}{Softmax} &
  Softmax \\ \hline
\multicolumn{1}{|l|}{\textbf{\begin{tabular}[c]{@{}l@{}}FC units\\ (Descending)\end{tabular}}} &
  \multicolumn{1}{c|}{512,   64, 8} &
  \multicolumn{1}{c|}{512,   64, 8} &
  \multicolumn{1}{c|}{512,   64, 8} &
  \multicolumn{1}{c|}{512, 64, 8} &
  512,   64, 8 \\ \hline
\multicolumn{6}{|c|}{\textbf{\begin{tabular}[c]{@{}c@{}} \\ Output sizes after joint processing blocks \footnotemark[2] \\ \end{tabular}}} \\ \hline
\multicolumn{1}{|l|}{\textbf{Module 1}} &
  \multicolumn{1}{c|}{4x24x32x24} &
  \multicolumn{1}{c|}{4x24x32x24} &
  \multicolumn{1}{c|}{4x24x32x24} &
  \multicolumn{1}{c|}{8x96x128x72} &
  8x96x128x72 \\ \hline
\multicolumn{1}{|l|}{\textbf{Module 2}} &
  \multicolumn{1}{c|}{2x12x16x48} &
  \multicolumn{1}{c|}{2x12x16x48} &
  \multicolumn{1}{c|}{2x12x16x48} &
  \multicolumn{1}{c|}{4x48x64x144} &
  4x48x64x144 \\ \hline
\multicolumn{1}{|l|}{\textbf{Module 3}} &
  \multicolumn{1}{c|}{1x6x8x96} &
  \multicolumn{1}{c|}{1x6x8x96} &
  \multicolumn{1}{c|}{1x6x8x96} &
  \multicolumn{1}{c|}{2x24x32x288} &
  2x24x32x288 \\ \hline
\multicolumn{1}{|l|}{\textbf{Module 4}} &
  \multicolumn{1}{c|}{1x3x4x192} &
  \multicolumn{1}{c|}{1x3x4x192} &
  \multicolumn{1}{c|}{1x3x4x192} &
  \multicolumn{1}{c|}{1x12x16x576} &
  1x12x16x576 \\ \hline
\multicolumn{1}{|l|}{\textbf{Module 5}} &
  \multicolumn{1}{c|}{-} &
  \multicolumn{1}{c|}{-} &
  \multicolumn{1}{c|}{-} &
  \multicolumn{1}{c|}{1x6x8x576} &
  1x6x8x1152 \\ \hline
\multicolumn{1}{|l|}{\textbf{Module 6}} &
  \multicolumn{1}{c|}{-} &
  \multicolumn{1}{c|}{-} &
  \multicolumn{1}{c|}{-} &
  \multicolumn{1}{c|}{1x3x4x384} &
  - \\ \hline
\end{tabular}%
}
\end{table}

\subsection{Model training}

All models were trained using the categorical cross-entropy loss and optimized using stochastic gradient descent (SGD). The number of epochs and batch size were set to 85 and 8, respectively. Training was set to reduce its learning rate by a factor of 0.5 if validation loss plateaus or peaks, and finally stop if no notable learning is achieved. This prevents overfitting and allows for early restoration of the best checkpoints. Each model is trained and evaluated $n=4$ times corresponding to different random seeds, and averaged. By using the averages, we present an accurate representation of each models’ predictive capability. 

\begin{table}[h]
\fontsize{17}{20}\selectfont
\centering
\caption{Model hyperparameters and other details}
\label{tab}
\resizebox{\textwidth}{!}{
\begin{tabular}{|l|l|l|l|l|l|l|l|l|}
\hline
\textbf{Models} &
  \begin{tabular}[c]{@{}l@{}}Baseline\end{tabular} &
  \begin{tabular}[c]{@{}l@{}}CIv3D\_BiLSTM\ \end{tabular} &
  \begin{tabular}[c]{@{}l@{}}CIv3D\_MHA\ \end{tabular} &
  \multicolumn{1}{c|}{SRS} &
  \multicolumn{1}{c|}{CRS} \\ \hline
\textbf{Learning Rate(s)}                                                        & \multicolumn{1}{c|}{0.0005} & \multicolumn{1}{c|}{0.001} & \multicolumn{1}{c|}{0.001}  & \multicolumn{1}{c|}{0.001} & \multicolumn{1}{c|}{0.001}    \\ \hline
\textbf{\begin{tabular}[c]{@{}l@{}}Parameters \\ (feature sharing)\end{tabular}} &
  \multicolumn{1}{c|}{11,315,848} &
  \multicolumn{1}{c|}{13,243,712} &
  \multicolumn{1}{c|}{15,554,112} &
  \multicolumn{1}{c|}{9,671,872} &
  \multicolumn{1}{c|}{22,927,824} \\ \hline
\textbf{\begin{tabular}[c]{@{}l@{}}Parameters \\ (standalone)\end{tabular}} &
  \multicolumn{1}{c|}{11,348,728} &
  \multicolumn{1}{c|}{16,623,752} &
  \multicolumn{1}{c|}{19,044,744} &
  \multicolumn{1}{c|}{9,670,024} &
  \multicolumn{1}{c|}{22,483,944} \\ \hline
\textbf{\begin{tabular}[c]{@{}l@{}}FLOPS \\ (feature sharing)\end{tabular}} &
  \multicolumn{1}{c|}{$22.63 \times 10^6$} &
  \multicolumn{1}{c|}{$30.67 \times 10^6$} &
  \multicolumn{1}{c|}{$31.10 \times 10^6$} &
  \multicolumn{1}{c|}{$19.34 \times 10^6$} &
  \multicolumn{1}{c|}{$45.85 \times 10^6$} \\ \hline
\textbf{\begin{tabular}[c]{@{}l@{}}FLOPS \\ (standalone)\end{tabular}} &
  \multicolumn{1}{c|}{$22.69 \times 10^6$} &
  \multicolumn{1}{c|}{$37.42 \times 10^6$} &
  \multicolumn{1}{c|}{$38.07 \times 10^6$} &
  \multicolumn{1}{c|}{$19.34 \times 10^6$} &
  \multicolumn{1}{c|}{$44.96 \times 10^6$} \\ \hline  
  \end{tabular} %
}\\
\end{table}

\subsection{Metrics} The most popular metric used in supervised classification is the accuracy. However, we evaluate all the models presented here on several metrics, including accuracy, average precision (AP), F1 Score, and area-under-the-ROC-curve (AUC), where ROC is the receiver operating characteristic. The plots depicting these metrics are the confusion matrices, precision-recall plots and ROC plots. Altogether, these metrics give a holistic view of each models’ performance. 


\footnotetext{ filter/unit size n in joint processors (subfigures \ref{inceptv3D} and \ref{simp}) is based on filter size after stream combination (i.e., adding or concatenation) and uses fixed multiplier k = 1.5.}
\footnotetext[2]{ Striding through temporal dimension produced more compact feature representations and lessened overall parameter count.}

\section{Results}\label{sec4}

\subsection{Model comparison}
The best results were obtained on the singly-reinforced stream model with an average accuracy of 81.96±2.71\%.\ The averaged performances of all the \emph{feature sharing} dual stream models are tabulated in Table \ref{AverPerfs}. The full performances for all seeds can be found in appendix \ref{secA1}. 

\begin{table}[ht] 
\centering
\caption{All models' performances across metrics}
\label{AverPerfs}
\resizebox{\textwidth}{!}{
\begin{tabular}{@{}cccclll@{}}
\toprule
\multirow{2}{*}{\textbf{MODELS}} &
  \multicolumn{5}{c}{\textbf{METRICS}} \\ \cmidrule(l){2-6} 
   &
  \multicolumn{2}{c}{AUC} &
  \multicolumn{1}{c}{\multirow{2}{*}{\begin{tabular}[c]{@{}c@{}}Average \\   Precision\end{tabular}}} &
  \multicolumn{1}{c}{\multirow{2}{*}{F1-Score}} &
  \multicolumn{1}{c}{\multirow{2}{*}{Accuracy}} \\ \cmidrule(lr){2-3}
 &
  \multicolumn{1}{c}{micro (m)} &
  \multicolumn{1}{c}{macro (M)} &
  \multicolumn{1}{c}{} &
  \multicolumn{1}{c}{} &
  \multicolumn{1}{c}{} \\ \midrule
\multirow{1}{*}{Baseline} &

  {0.8785} &
  {0.9197} &
  {0.592} &
  {0.5515} &
  {74.93} \\ \midrule
\multirow{1}{*}{\begin{tabular}[c]{@{}c@{}}CIv3D\_BiLSTM\ \end{tabular}} &
  
  {0.9553} &
  {0.9648} &
  {0.789} &
  {0.6540} &
  {77.08} \\ \midrule
\multirow{1}{*}{\begin{tabular}[c]{@{}c@{}}CIv3D\_MHA\ \end{tabular}}  
 &
  {0.9505} &
  {0.9532} &
  {0.770} &
  {0.6779} &
  {77.99} \\ \midrule
\multirow{1}{*}{\begin{tabular}[c]{@{}c@{}}SRS\end{tabular}} &
  
  {0.9586} &
  {0.9716} &
  {0.791} &
  {0.6863} &
  {{\textit{\textbf{81.96}}}} \\ \midrule
\multirow{1}{*}{\begin{tabular}[c]{@{}c@{}}CRS\end{tabular}} &
  
  {0.9340} &
  {0.9601} &
  {0.728} &
  {0.5775} &
  {78.49} \\ \bottomrule 
\end{tabular}
}
\end{table}
 
\subsection{Ensembles} \label{ensembles}
The ensembles were created by averaging the results of the models at inference. Due to the gap in performance, most ensembles between models did not show any improvements over the SRS model. 
However, by evaluating both the validation and test results, the best and second-best performing seeds of the SRS model were ensemled. For intra-model ensembles, the test results yielded 82.37\%\ based on model picks via evaluating of the validation data. This further increases to an accuracy of 85.9\%\ based on the model picks using the test data itself. By evaluating the test results (see appendix \ref{secA1}), the best inter-model ensemble was observed between the SRS and CIv3D\_MHA models and achieved 86.47\%.\ Further ensembles between models are shown in Table \ref{tab3}. The confusion matrices and ROC plots for the ensembles can be found in appendix \ref{secA2}.

\begin{table}[ht]
\centering
\caption{Result of ensembles (based on test groundtruth)}
\label{tab3}
\resizebox{\textwidth}{!}{%
\begin{tabular}{llllll}
\hline
\multirow{2}{*}{\textbf{MODEL ENSEMBLES}} & \multicolumn{3}{l} \ \textbf{METRICS}\ \\ \cline{2-6} 
 & \textbf{mAUC} & \textbf{MAUC} & \textbf{A}P & \textbf{F1 Score} & \textbf{Accuracy (\%)}\ \\ \hline
\begin{tabular}[c]{@{}l@{}}\ CIv3D\_ MHA\ + \\ \ CIv3D\_\ BiLSTM\ \end{tabular} & 0.9788 & 0.9730 & 0.882 & 0.7888 & 83.22 \\ \hline
\begin{tabular}[c]{@{}l@{}}\ SRS\ + \\ \ CIv3D\_MHA\ \end{tabular} & 0.9751 & 0.9760 & 0.869 & 0.7785 & \textbf{86.47} \\ \hline
\begin{tabular}[c]{@{}l@{}}\ SRS\ + \\ \ CIv3D\_BiLSTM \ \end{tabular} & 0.9767 & 0.9770 & 0.874 & 0.7808 & 85.72 \\ \hline
All three ensembled & 0.9789 & 0.9770 & 0.885 & 0.7976 & 86.24 \\ \hline
\end{tabular}%
}
\footnotetext{\textbf{NOTE:} mAUC - micro AUC, MAUC - macro AUC, AP - Average Precision}

\end{table}

\subsection{Ablation study} 
\subsubsection{The case for \emph{feature sharing}}
Here, the results of the models and their non \emph{feature sharing} variants are presented. The variants were trained and tested on the same dataset, and under the same conditions as those with joint processing. The averaged results across all metrics are tabulated (table \ref{tab2}). It can be clearly observed that the for each architecture pair, the \emph{feature sharing} models performs better than their standalone forms.
 
\begin{table}
\fontsize{17}{20}\selectfont
\centering
\caption{Summary of the feature-sharing and standalone stream networks}
\label{tab2}
\resizebox{\textwidth}{!}{
\begin{tabular}{lllllll}
\hline
\multicolumn{1}{c}{\textbf{Models}} &
  \multicolumn{1}{c}{\textbf{Stream kind}} &
  \multicolumn{1}{c}{\textbf{mAUC}} &
  \multicolumn{1}{c}{\textbf{MAUC}} &
  \multicolumn{1}{c}{\textbf{AP}} &
  \multicolumn{1}{c}{\textbf{F1 Score}} &
  \multicolumn{1}{c}{\textbf{Accuracy (\%)}}\ \\ \hline
\multirow{2}{*}{Baseline}        & sharing    & 0.878±0.013 & 0.920±0.004 & 0.592±0.017 & 0.552±0.026 & 74.93±2.68  \\ \cline{2-7} 
                                 & standalone & 0.815±0.020 & 0.916±0.010 & 0.562±0.024 & 0.483±0.016 & 70.28±1.64  \\ \hline
\multirow{2}{*}{CIv3DBiLSTM}     & sharing    & 0.955±0.012 & 0.965±0.003 & 0.789±0.045 & 0.654±0.074 & 77.08±1.77  \\ \cline{2-7} 
                                 & standalone & 0.910±0.015 & 0.955±0.004 & 0.668±0.051 & 0.537±0.026 & 76.95±2.02  \\ \hline
\multirow{2}{*}{CIv3DMHA}        & sharing    & 0.951±0.017 & 0.953±0.017 & 0.770±0.060 & 0.678±0.074 & 77.99±3.50  \\ \cline{2-7} 
                                 & standalone & 0.896±0.018 & 0.938±0.007 & 0.635±0.036 & 0.564±0.034 & 73.10±2.44  \\ \hline
\multirow{2}{*}{SRS} & sharing    & 0.959±0.010 & 0.972±0.006 & 0.791±0.042 & 0.686±0.050 & 81.96±2.71  \\ \cline{2-7} 
                                 & standalone & 0.900±0.029 & 0.931±0.009 & 0.632±0.064 & 0.666±0.048 & 71.07±1.59  \\ \hline
\multirow{2}{*}{CRS} & sharing    & 0.934±0.006 & 0.960±0.004 & 0.728±0.015 & 0.578±0.018 & 78.49±0.945 \\ \cline{2-7} 
                                 & standalone & 0.868±0.016 & 0.921±0.008 & 0.562±0.025 & 0.523±0.019 & 67.95±1.804 \\ \hline
\end{tabular}%
}
\end{table}

\subsubsection{The case for \emph{nightification}}
To justify the choice of \emph{nightified} spatiotemporal (ST) clips in the image stream, further experiments were conducted for both raw rgb input and grayscale input. This were carried out on the baseline model and the previously ascertained best performing models from section \ref{ensembles}. These models were trained and testsed in the same rigorous manner as the core paper models. 
The results show that \emph{nightified} ST input has higher accuracy than both grayscale and raw video ST inputs for most models, the only exception being the baseline model. Those using grayscale cuboids seemed to initially perform well just observing the AUCs and average precision however all their accuracies were subpar to the \emph{nightified} cuboids. Observations show that this was due to greater misclassification between visually similar behaviours (such as \emph{micromovement} and \emph{rest}), indicative of the fact that the grayscale modality did not possess enough information for these deep models to sufficiently distinguish between . A similar narrative was observed in the raw video inputs, though we argue that in this case that the drop in performance (albeit small) was due to the lack of standardization. The results are presented in table \ref{varyDataForms}.

\begin{table}[ht]
\caption{Results on grayscale, raw RGB and \emph{nightified} data}
\label{varyDataForms}
\begin{tabular}{|l|lll|lll|lll|}
\hline
\textbf{Model}         & \multicolumn{3}{l|}{\textbf{Baseline}}                                        & \multicolumn{3}{l|}{\textbf{CIv3D\_MHA}}                                 & \multicolumn{3}{l|}{\textbf{SRS}}                                        \\ \hline
\textbf{Input kind?}   & \multicolumn{1}{l|}{GS}    & \multicolumn{1}{l|}{R}              & \textit{N} & \multicolumn{1}{l|}{GS}    & \multicolumn{1}{l|}{R}     & \textit{N}     & \multicolumn{1}{l|}{GS}    & \multicolumn{1}{l|}{R}     & \textit{N}     \\ \hline
\textbf{mAUC}          & \multicolumn{1}{l|}{0.873} & \multicolumn{1}{l|}{0.869}          & 0.879      & \multicolumn{1}{l|}{0.910} & \multicolumn{1}{l|}{0.921} & 0.950          & \multicolumn{1}{l|}{0.962} & \multicolumn{1}{l|}{0.938} & 0.959          \\ \hline
\textbf{MAUC}          & \multicolumn{1}{l|}{0.910} & \multicolumn{1}{l|}{0.914}          & 0.920      & \multicolumn{1}{l|}{0.929} & \multicolumn{1}{l|}{0.943} & 0.953          & \multicolumn{1}{l|}{0.972} & \multicolumn{1}{l|}{0.953} & 0.972          \\ \hline
\textbf{AP}            & \multicolumn{1}{l|}{0.545} & \multicolumn{1}{l|}{0.577}          & 0.592      & \multicolumn{1}{l|}{0.647} & \multicolumn{1}{l|}{0.673} & 0.770          & \multicolumn{1}{l|}{0.810} & \multicolumn{1}{l|}{0.687} & 0.791          \\ \hline
\textbf{F1 Score}      & \multicolumn{1}{l|}{0.542} & \multicolumn{1}{l|}{0.544}          & 0.552      & \multicolumn{1}{l|}{0.564} & \multicolumn{1}{l|}{0.597} & 0.678          & \multicolumn{1}{l|}{0.733} & \multicolumn{1}{l|}{0.569} & 0.686          \\ \hline
\textbf{Accuracy (\%)} & \multicolumn{1}{l|}{73.37} & \multicolumn{1}{l|}{\textbf{76.57}} & 74.93      & \multicolumn{1}{l|}{73.70} & \multicolumn{1}{l|}{77.16} & \textbf{77.99} & \multicolumn{1}{l|}{79.24} & \multicolumn{1}{l|}{77.87} & \textbf{81.96} \\ \hline
\end{tabular}
\footnotetext{Where GS – grayscale, R- raw RGB, \emph{N} – nightified data }
\end{table}

\subsubsection{Varying of temporal length}
The temporal length refers to the number of frames that make up each clip. As previously stated, all architectures were designed for a temporal length \emph{T=8}, corresponding to 1.33 seconds. Further experiments are performed here by varying the preset \emph{T} value. The new temporal lengths chosen were (i.e.\emph{T=4}) and (i.e.\emph{T=16}). These experiments were only carried out on the baseline and SRS models, and conducted in the same rigorous manner as the initial runs. Besides changing the input shape, the temporal cropping (refer to section \ref{srs_crs}) in the SRS architecture was also slightly modified. Same as the new T values, this feature was halved and doubled respectively for \emph{T=4} and \emph{T=16}. Hence, there was no change to the architectural complexity. For the baseline model, its complexity only increased, slightly, for \emph{T=16}. The results after averaging the results for various seeds are shown in table \ref{tempComparisons}.

\begin{table}[h]
\caption{Results for varied temporal lengths}
\label{tempComparisons}
\centering
\resizebox{\textwidth}{!}{%
\begin{tabular}{|l|c|c|c|c|c|c|}
\hline
 &
  \textbf{Temporal length(s)} &
  \textbf{mAUC} &
  \textbf{MAUC} &
  \multicolumn{1}{l|}{\textbf{AP}} &
  \multicolumn{1}{l|}{\textbf{F1 Score}} &
  \multicolumn{1}{l|}{\textbf{Accuracy (\%)}} \\ \hline
\multirow{3}{*}{\textbf{Baseline}} & 4  & 0.8532 & 0.7430 & 0.411 & 0.3807 & 57.03          \\ \cline{2-7} 
                                   & 8  & 0.8785 & 0.9197 & 0.592 & 0.5515 & \textbf{74.93} \\ \cline{2-7} 
                                   & 16 & 0.8314 & 0.7364 & 0.365 & 0.3485 & 47.63          \\ \hline
\multirow{3}{*}{\textbf{SRS}}      & 4  & 0.9135 & 0.9575 & 0.629 & 0.5759 & 69.72          \\ \cline{2-7} 
                                   & 8  & 0.9586 & 0.9716 & 0.791 & 0.6863 & \textbf{81.96} \\ \cline{2-7} 
                                   & 16 & 0.9155 & 0.9298 & 0.658 & 0.6071 & 67.43          \\ \hline
\end{tabular}
}
\footnotetext{\textbf{NOTE:} mAUC - micro AUC, MAUC - macro AUC, AP - Average Precision}
\end{table}

The results show that the preset \emph{T=8} was optimum as the accuracies obtain in the new experiments were not upto par. 

\subsection{Other datasets}
\subsubsection{SCORHE}
Further experiments were conducted by applying the pretrained versions of the top three seeds (from all models) to a new home-caged mouse dataset. As previously shown in Section \ref{ensembles}, the top performing seeds occur in CIv3D\_BiLSTM,\ CIv3D\_MHA\ and SRS models. The dataset of choice is another singly housed-mouse data by SCORHE \citep{salem2015scorhe}. Although 13 unique annotations were originally present (see graph in Appendix \ref{secA3}), these were refined to 8 classes by removing samples with ambiguous classes (such as \emph{behav\_ignore, behav\_other}), removing samples having extremely low class occurrence (such as \emph{discrepancy, rotating}), and merging the supported and unsupported \emph{rearing} classes. 

The recordings in the SCORHE home cage were done from multiple points as no singular viewpoint provides a clear view with occlusions. To address this, the viewpoints from opposite ends of SCORHE were shaped as 128x64 frames and stacked into a singular $128\times128$ frame. The same was also done for the optical flow data. No frame skips were used here to ensure ample training and testing data was available. Data samples for the SCORHE dataset are available in Appendix \ref{secA4}.

\begin{figure}[ht]
    \centering
    \subfigure[Feature sharing streams]{\label{FS_SCORHE}\includegraphics[width=\textwidth, trim={2.5cm 4.5cm 2.5cm 5cm}, clip, page=2]{figure_3ModsSCORHE_micro_ROC_PR_2.pdf}}
    
    \subfigure[Standalone streams]{\label{StdAlone_SCORHE}\includegraphics[width=\textwidth, trim={2.5cm 4.5cm 2.5cm 5cm}, clip,  page=1]{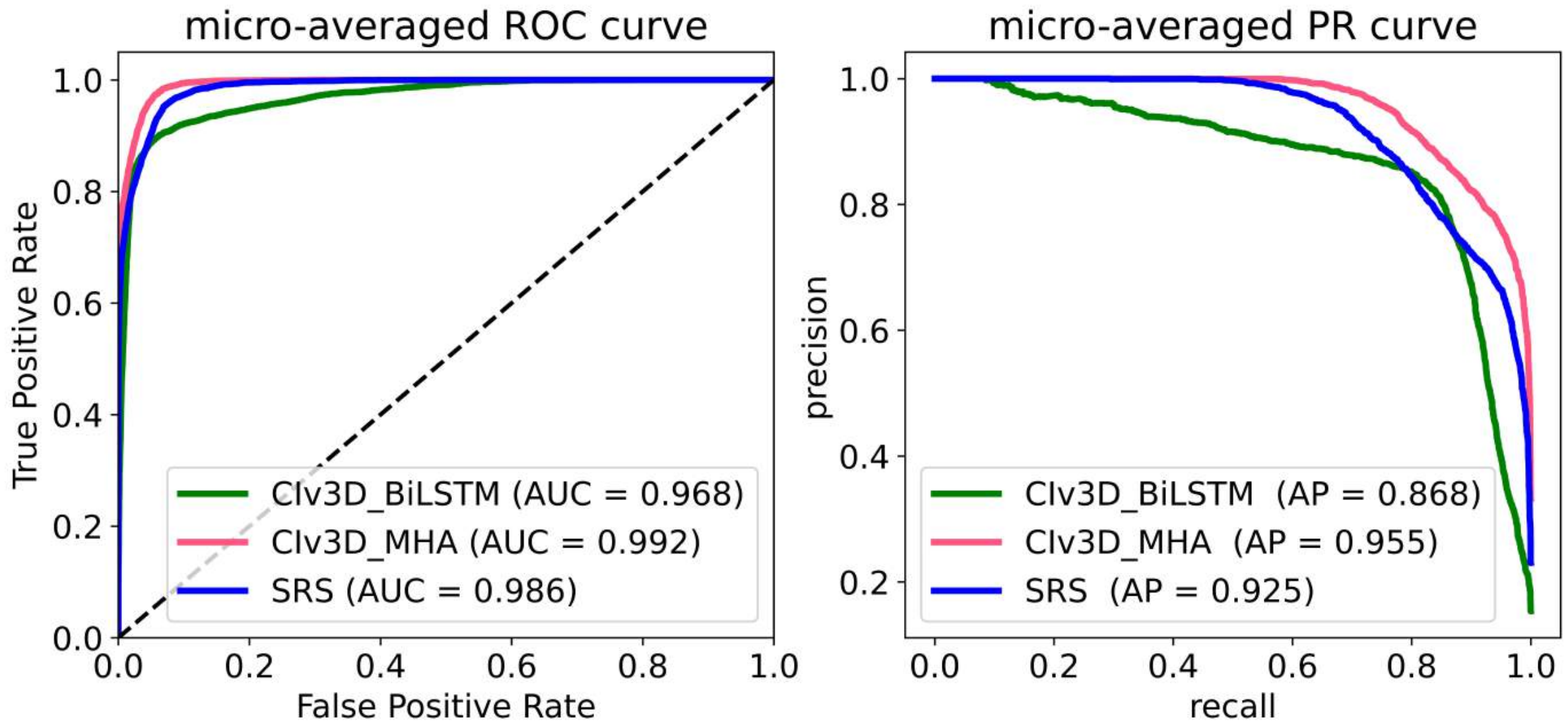}}
    \caption{ROC and PR curves on SCORHE dataset.}
    \label{scorheROCPR}
\end{figure}

For the training, the previous FC layers were changed for new ones. All other training parameters remained the same, save the learning rate which was halved to 0.0005. The resulting receiver operating characteristics (ROC) and precision-recall (PR) curves are shown in figure \ref{scorheROCPR}. The accuracies achieved on the SCORHE dataset by the \emph{feature sharing} CIv3D\_BiLSTM\, CIv3D\_MHA\ and SRS were 80.51\%,\ 79.88\%\ and 79.13\%\ respectively. Their non \emph{feature sharing} variants achieved 72.18\%,\ 77.95\%\ and 70.83\%\ respectively. 

A few observations were made on the \emph{feature sharing} models. The CIv3D\_BiLSTM\ and CIv3D\_MHA\ were good at reinforcing previously learnt spatiotemporal representations to this complex home cage for similar behaviours. However, despite having lower accuracy, SRS performed better in both learning old classes and balancing predictions to learn totally new class, \emph{climbing}. This is proven by its class accuracy across the different confusion matrices; while CIv3D\_BiLSTM\ and CIv3D\_MHA\ achieved 22.34\%\ and 33.68\%\ respectively, the SRS model achieved 53.61\%.\

\subsubsection{UCF101}
Finally, the \emph{feature sharing} and standalone forms of the best performing model (i.e. SRS) were applied to transfer learning on a popular, more challenging activity recognition dataset; the UCF101 \citep{soomro2012ucf101}. This dataset contains 13,320 clips of 101 activity classes totalling over 27 hours. In a similar manner as before, the models (pretrained on the MIT mouse dataset) are utilized, having new fully-connected layers and the learning rate reduced to 0.0005.  After shaping data into 8-frame cuboids of $128\times128$, a train/validation/test split of 0.64/0.16/0.20 was applied. This experiment was done purely as a cross-domain investigation into the effectiveness of \emph{feature sharing} so no further preprocessing was carried out on either the RGB or optical flow data. As shown in table \ref{ucf101results}, the accuracy of the pretrained SRS yet again bests its non \emph{feature sharing} counterpart across the board. In addition, the top-5 accuracy of the SRS, without pretraining, is seen reaching very high accuracies. Comparison with SOTA results however is not feasible since data split was done differently.

\begin{table}[ht]
\centering
\caption{Results on UCF101 dataset}
\label{ucf101results}
\begin{tabular}{|l|ccc|}
\hline
\textbf{Model}               & \multicolumn{3}{c|}{SRS}  
\\ \hline

\textbf{Model kind?} &
  \multicolumn{1}{c|}{\begin{tabular}[c]{@{}c@{}}Standalone (p)\end{tabular}} &
  \multicolumn{1}{c|}{\begin{tabular}[c]{@{}c@{}}Feature sharing (p)\end{tabular}} &
  \multicolumn{1}{l|}{Feature sharing (np)} \\ \hline
\textbf{New parameter size}  & \multicolumn{1}{c|}{12,100,005} & \multicolumn{1}{c|}{12,101,853} & 12,101,853 \\ \hline
\textbf{mAUC}                & \multicolumn{1}{c|}{0.989}      & \multicolumn{1}{c|}{0.997}      & 0.998      \\ \hline
\textbf{MAUC}                & \multicolumn{1}{c|}{0.985}      & \multicolumn{1}{c|}{0.994}      & 0.997      \\ \hline
\textbf{AP}                  & \multicolumn{1}{c|}{0.813}      & \multicolumn{1}{c|}{0.926}      & 0.957      \\ \hline
\textbf{F1 Score}            & \multicolumn{1}{c|}{0.720}      & \multicolumn{1}{c|}{0.835}      & 0.875      \\ \hline
\textbf{Top-1 accuracy (\%)}\ & \multicolumn{1}{c|}{73.82}      & \multicolumn{1}{c|}{86.02}      & 90.23      \\ \hline
\textbf{Top-5 accuracy (\%)}\ & \multicolumn{1}{c|}{90.05}      & \multicolumn{1}{c|}{95.25}      & 97.23      \\ \hline
\end{tabular}
\footnotetext{\textbf{NOTE:} p - pretrained on mouse dataset, np - no pretraining}
\end{table}

\section{Conclusion}\label{sec5}

Generally, it was observed that the more dynamic behaviours were better captured by all the models. The performance lag in all the  models was mainly due to  misclassifications amongst the \emph{resting}, \emph{grooming} and \emph{micromovement} behaviours. These behaviours are quite closely related; during \emph{grooming}, the mouse is mostly stationary albeit the motion of its forelimbs and when resting, the mouse is immobile. \emph{Micromovement} are very small-scale motions. Hence, it is most likely that the 1.33 second windows of \emph{T=8} cuboids cannot not capture the full range of motions that will allow the models better distinguish between these classes. Nonetheless, these 'misclassifications' are also indicative of a similitude in the temporal pattern needed to perform certain tasks and may be subject to further interpretation by the subject experts. Further experiments in the ablation study also showed that for time windows lower or higher the  1.33 second window, the performance of the models degrade. Thus these ST clips (especially for \emph{T=16}) may require more specialized model designs to work with the \emph{feature sharing} framework.

The step up in performance between the \emph{feature sharing} and standalone baseline models lends credence to the effectiveness of combined streams; by simply summing parallel outputs from both streams and processing with a dense-dropout pair, we observe over 4\%\ improvement in averaged accuracy. This observation was further proven in subsequent networks utilizing algorithms such as bidirectional LSTMs and self-attention mechanisms. Though the CIv3D\_BiLSTM\ model was only marginally better in terms of accuracy, it bested its non \emph{feature sharing} variant in all other metrics. Similarly, we observe a notable boost across all the metrics of the other models, especially so in the purely 3D Inception-based networks (SRS and CRS), both having over 10\%\ improvement in accuracy alone. Future research will also consider unsupervised detection of behaviours and welfare concerns in the home cage, and if the unique \emph{feature sharing} approach will impact multi-stream models in this learning domain.

\backmatter
\section*{Acknowledgments}
This research was funded by the National Centre for the Replacement, Refinement and Reduction of Animals in Research (NC/T002050/1).

\bibliography{Manuscript}


\newpage
\begin{appendices} %

\section{Full performance table }\label{secA1}

\begin{table}[h]
\centering
\caption{Full Performance table for feature sharing models}
\label{tab:FullPerfTab}
\resizebox{\textwidth}{!}{%
\begin{tabular}{lllllll}
\hline
\multicolumn{1}{c}{\multirow{3}{*}{MODELS}} &
  \multicolumn{1}{c}{\multirow{3}{*}{SEEDS}} &
  \multicolumn{5}{c}{METRICS} \\ \cline{3-7} 
\multicolumn{1}{c}{} &
  \multicolumn{1}{c}{} &
  \multicolumn{2}{c}{AUC} &
  \multicolumn{1}{c}{\multirow{2}{*}{AP}} &
  \multicolumn{1}{c}{\multirow{2}{*}{F1 Score}} &
  \multicolumn{1}{c}{\multirow{2}{*}{Accuracy (\%)}} \\ \cline{3-4}
\multicolumn{1}{c}{} &
  \multicolumn{1}{c}{} &
  \multicolumn{1}{c}{micro (m)} &
  \multicolumn{1}{c}{macro (M)} &
  \multicolumn{1}{c}{} &
  \multicolumn{1}{c}{} &
  \multicolumn{1}{c}{} \\ \hline
\multirow{5}{*}{Baseline} &
  A &
  0.89139 &
  0.92472 &
  0.602 &
  0.53434 &
  75.0425 \\ \cline{2-7} 
 &
  B &
  0.87151 &
  0.91915 &
  0.609 &
  0.57256 &
  77.4076 \\ \cline{2-7} 
 &
  C &
  0.86057 &
  0.91478 &
  0.564 &
  0.51816 &
  70.5459 \\ \cline{2-7} 
 &
  D &
  0.89047 &
  0.92020 &
  0.593 &
  0.58093 &
  76.7404 \\ \cline{2-7} 
 &
  \textbf{Average(s)} &
  \textbf{0.87849} &
  \textbf{0.91971} &
  \textbf{0.592} &
  \textbf{0.55150} &
  \textbf{74.9341} \\ \hline
\multirow{5}{*}{CIv3D\_\ BiLSTM} &
  A &
  0.93746 &
  0.   95331 &
  0.729 &
  0.55750 &
  76.3368 \\ \cline{2-7} 
 &
  B &
  0.92415 &
  0.95030 &
  0.720 &
  0.55634 &
  75.4044 \\ \cline{2-7} 
 &
  C &
  0.92848 &
  0.94623 &
  0.723 &
  0.53677 &
  75.7875 \\ \cline{2-7} 
 &
  D &
  0.91047 &
  0.95133 &
  0.729 &
  0.54228 &
  78.6451 \\ \cline{2-7} 
 &
  \textbf{Average(s)} &
  \textbf{0.92514} &
  \textbf{0.95029} &
  \textbf{0.725} &
  \textbf{0.54822} &
  \textbf{76.5435} \\ \hline
\multirow{5}{*}{CIv3D\_\ MHA} &
  A &
  0.92600 &
  0.94809 &
  0.692 &
  0.56606 &
  74.2224 \\ \cline{2-7} 
 &
  B &
  0.94715 &
  0.94632 &
  0.757 &
  0.68050 &
  76.2464 \\ \cline{2-7} 
 &
  C &
  0.97321 &
  0.97059 &
  0.860 &
  0.77441 &
  83.6366 \\ \cline{2-7} 
 &
  D &
  0.95560 &
  0.94758 &
  0.772 &
  0.69051 &
  77.8722 \\ \cline{2-7} 
 &
  \textbf{Average(s)} &
  \textbf{0.95049} &
  \textbf{0.95315} &
  \textbf{0.770} &
  \textbf{0.67787} &
  \textbf{77.9944} \\ \hline
\multirow{5}{*}{SRS} &
  A &
  0.96821 &
  0.97703 &
  0.839 &
  0.74434 &
  84.2392 \\ \cline{2-7} 
 &
  B &
  0.95825 &
  0.97383 &
  0.783 &
  0.65857 &
  79.4152 \\ \cline{2-7} 
 &
  C &
  0.94276 &
  0.96119 &
  0.727 &
  0.62004 &
  79.1211 \\ \cline{2-7} 
 &
  D &
  0.96502 &
  0.97416 &
  0.816 &
  0.72210 &
  85.0749 \\ \cline{2-7} 
 &
  \textbf{Average(s)} &
  \textbf{0.95856} &
  \textbf{0.97155} &
  \textbf{0.791} &
  \textbf{0.68626} &
  {\ul \textbf{81.9626}} \\ \hline
\multirow{5}{*}{CRS} &
  A &
  0.92582 &
  0.95732 &
  0.713 &
  0.56682 &
  77.2607 \\ \cline{2-7} 
 &
  B &
  0.93584 &
  0.96622 &
  0.728 &
  0.57181 &
  79.9323 \\ \cline{2-7} 
 &
  C &
  0.932344 &
  0.95750 &
  0.717 &
  0.56274 &
  78.0969 \\ \cline{2-7} 
 &
  D &
  0.94216 &
  0.95938 &
  0.752 &
  0.60866 &
  78.6730 \\ \cline{2-7} 
 &
  \textbf{Average(s)} &
  \textbf{0.93404} &
  \textbf{0.96011} &
  \textbf{0.728} &
  \textbf{0.57751} &
  \textbf{78.4907} \\ \hline
\end{tabular}%
}
\end{table}

\section{Additional performance plots}\label{secA2}

\begin{figure}[H]%
    \vspace*{-.75cm}
    \centering
    \hspace*{-1.75cm} 
    \includegraphics[scale=.55,page=8,trim={0.5cm 4.5cm 0 5cm},clip]{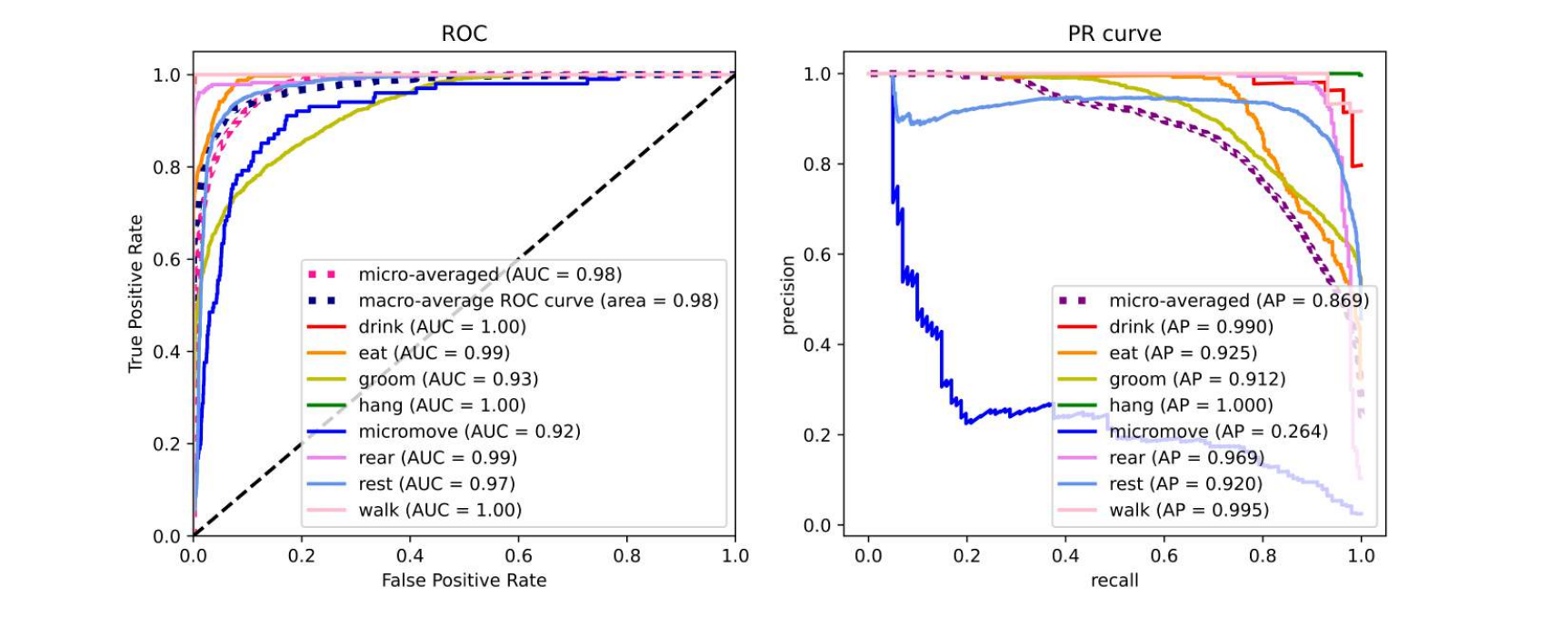}%
    \caption{ROC and PR plot for SRS + CIv3D\_MHA}%
    \label{Incept+CIv3D_MHAROC}%
\end{figure}%
\begin{figure}[H]%
    \vspace*{-.25cm}
    \centering
    \vspace*{-1cm} 
    \includegraphics[scale=.65,page=7,trim={0cm 6.5cm 0 6.25cm},clip]{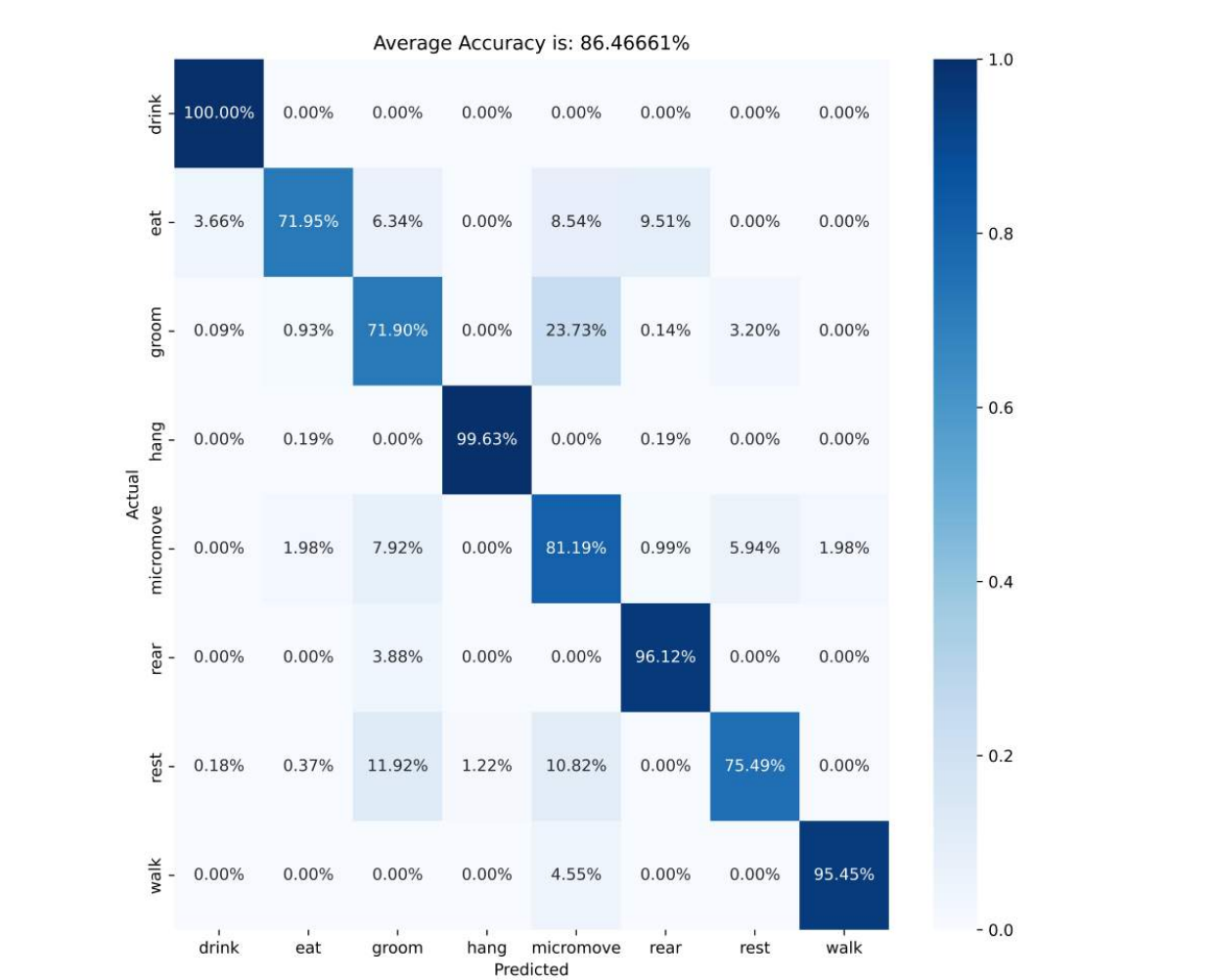}%
    \caption{Confusion Matrix for SRS + CIv3D\_MHA}%
    \label{cmInceptv1+CIv3D_MHA}%
\end{figure}

\begin{figure}[H]%
    \centering
    \hspace*{-1.75cm} 
    \includegraphics[scale=.55,page=6,trim={0.5cm 4.5cm 0 5cm},clip]{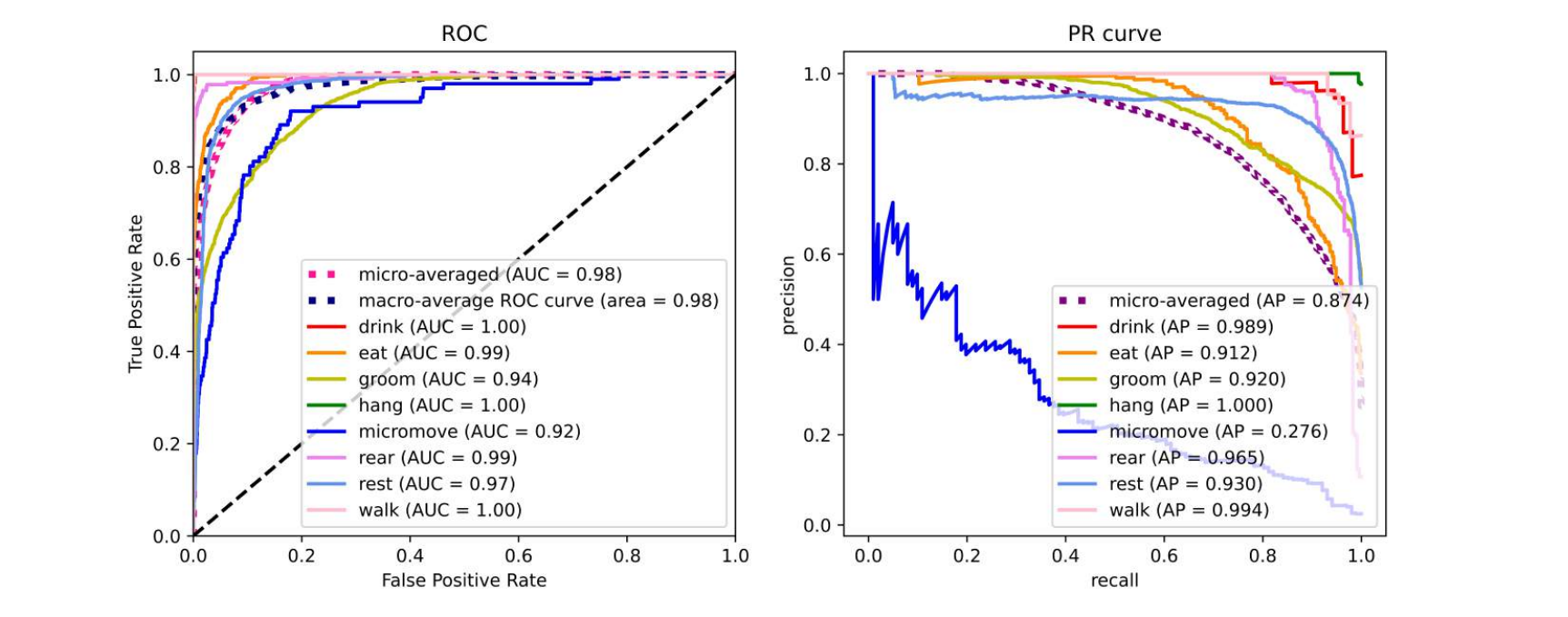}%
    \caption{ROC and PR plot for SRS + CIv3D\_BiLSTM}%
    \label{Incept+CIv3D_BiLSTMROC}%
\end{figure}%
\begin{figure}[H]%
    \centering
    \vspace*{-1cm} 
    \includegraphics[scale=.65,page=5,trim={0cm 6.5cm 0 6.25cm},clip]{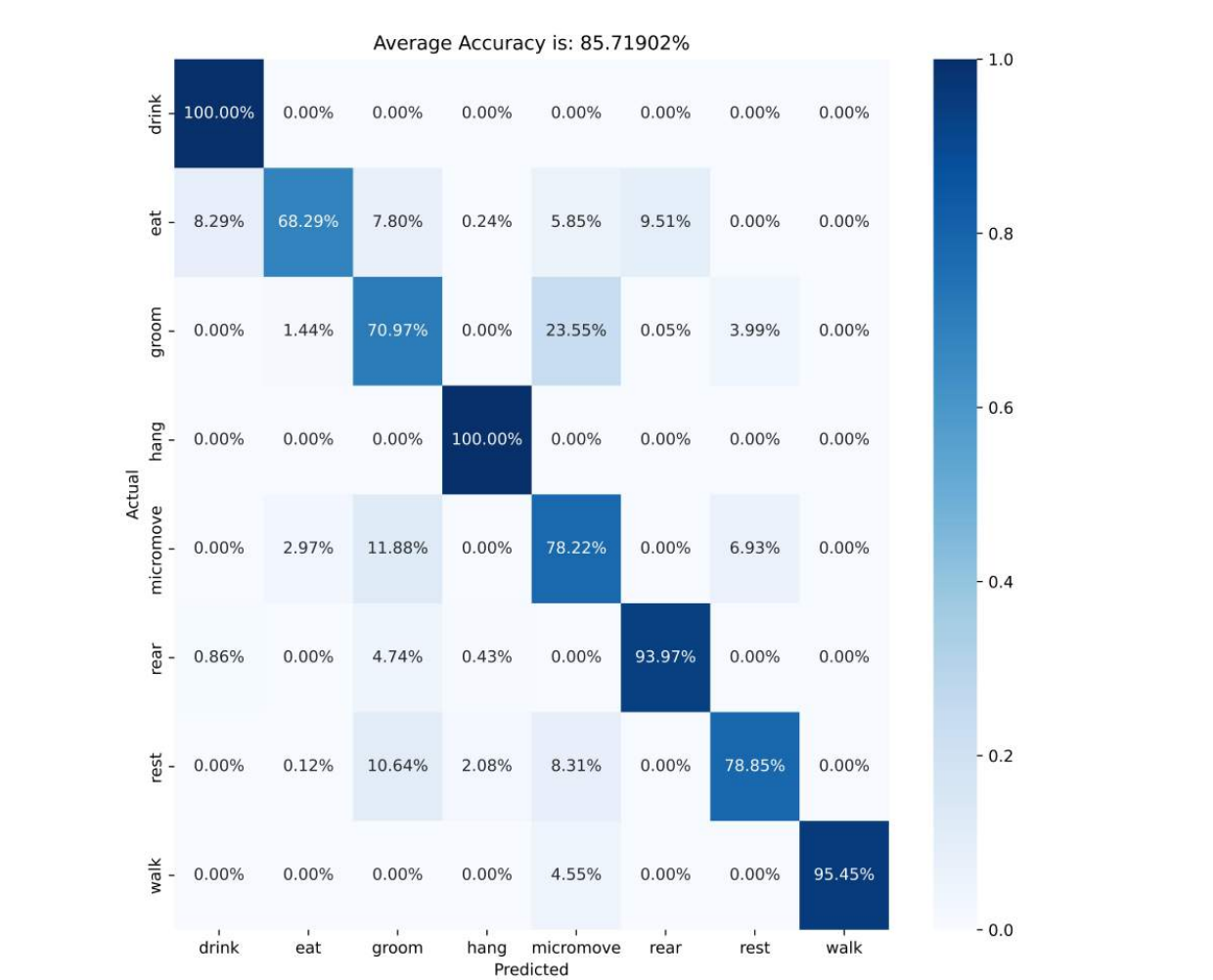}%
    \caption{Confusion Matrix for SRS + CIv3D\_BiLSTM}%
    \label{cmInceptv1+CIv3D_BiLSTM}%
\end{figure}

\begin{figure}[H]%
    \centering
    \hspace*{-1.75cm} 
    \includegraphics[scale=.55,page=2,trim={0.5cm 4.5cm 0 5cm},clip]{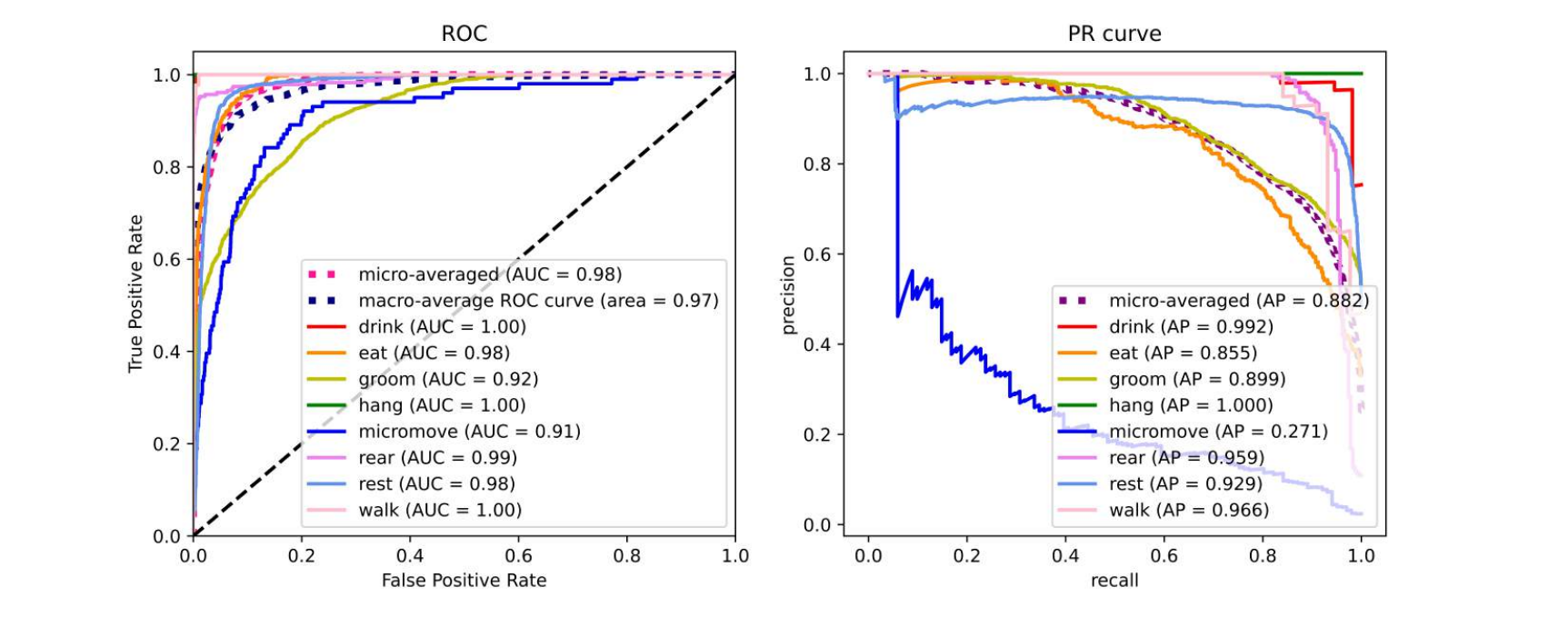}%
    \caption{ROC and PR plot for CIv3D\_BiLSTM + CIv3D\_MHA}%
    \label{CIv3D_BiLSTM+CIv3D_MHAROC}%
\end{figure}%
\begin{figure}[H]%
    \centering
    \vspace*{-1cm} 
    \includegraphics[scale=.65,page=1,trim={0cm 6.5cm 0 6.25cm},clip]{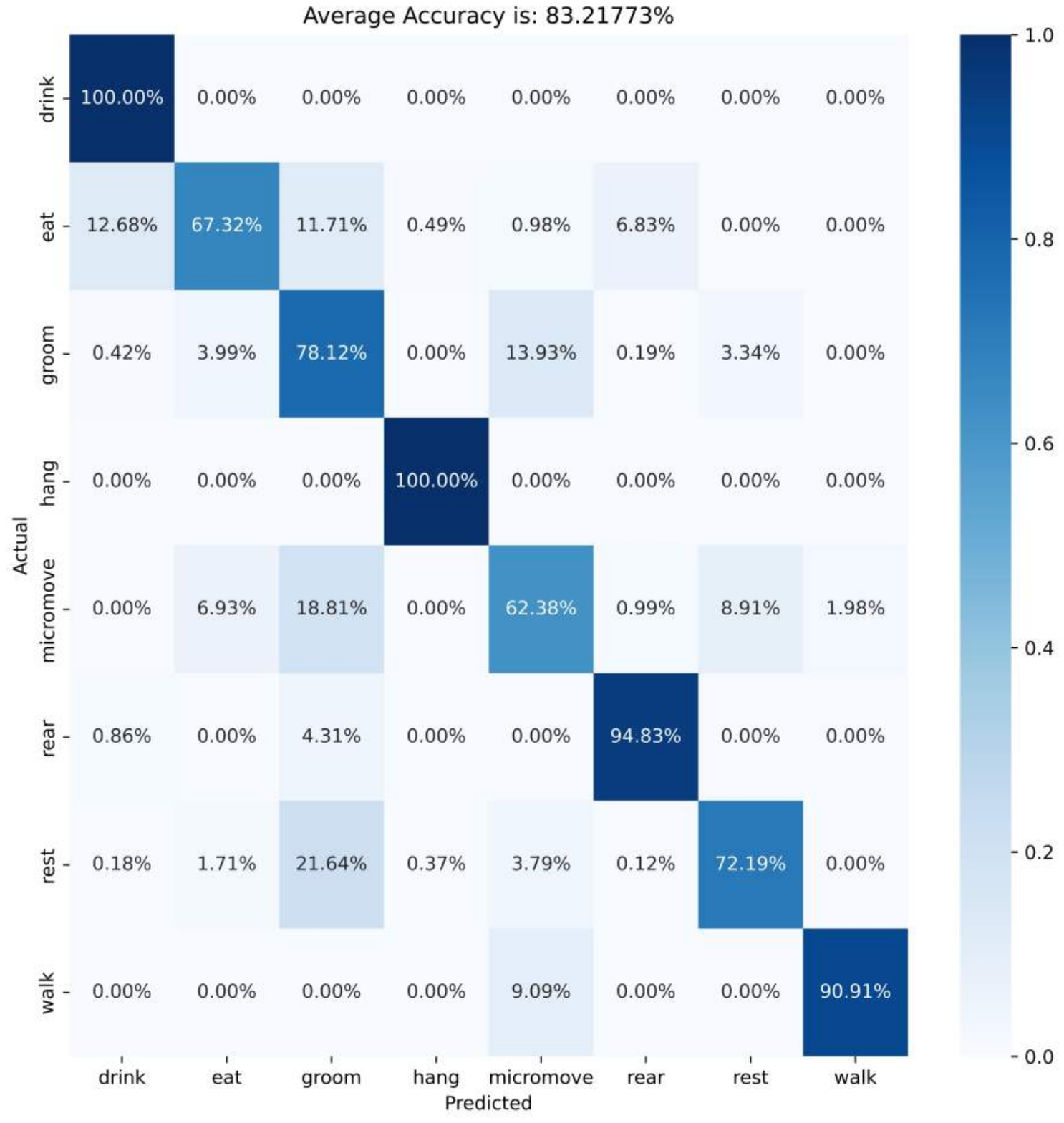}%
    \caption{Confusion Matrix for CIv3D\_BiLSTM + CIv3D\_MHA}%
    \label{cmCIv3D_BiLSTM+CIv3D_MHA}%
\end{figure}

\begin{figure}[H]%
    \centering
    \hspace*{-1.75cm} 
    \includegraphics[scale=.55,page=4,trim={0.5cm 4.5cm 0 5cm},clip]{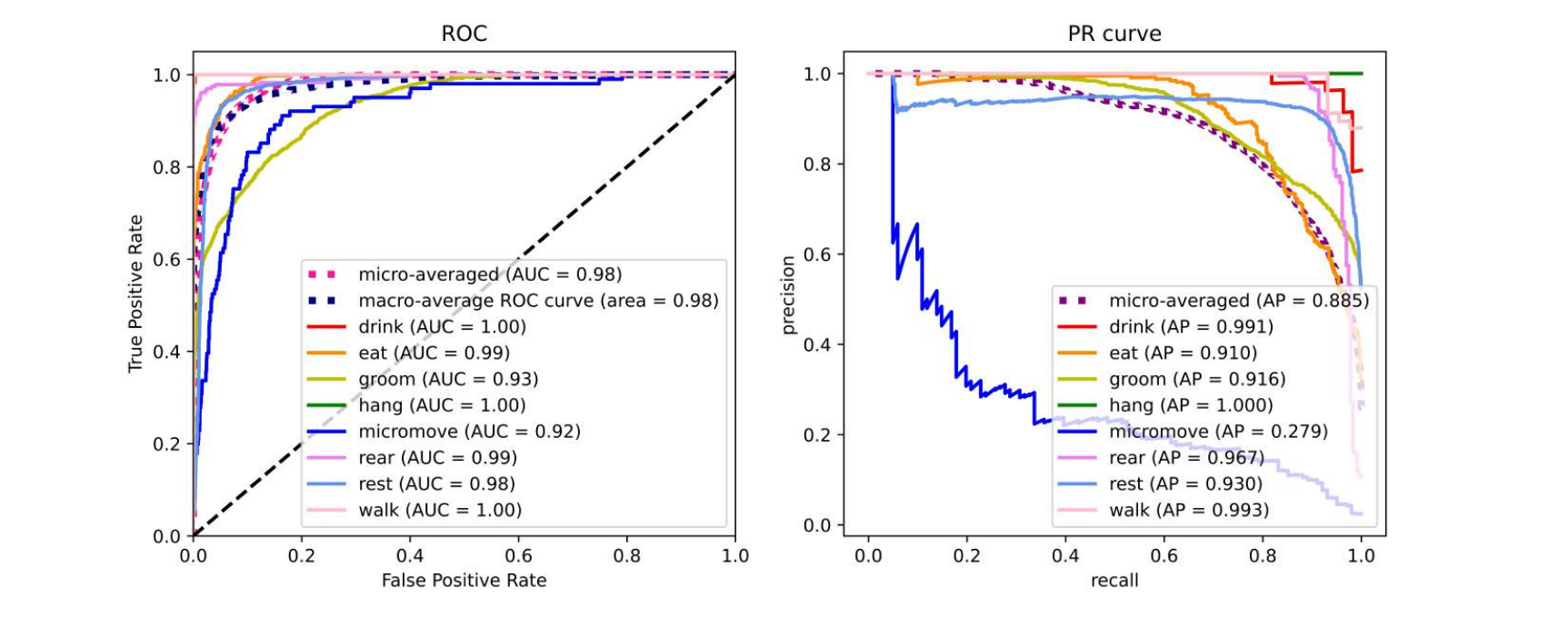}%
    \caption{ROC and PR plot for SRS + CIv3D\_BiLSTM  + CIv3D\_MHA}%
    \label{Incept+CIv3D_BiLSTM+CIv3D_MHAROC}%
\end{figure}%
\begin{figure}[H]%
    \centering
    \vspace*{-1.5cm} 
    \includegraphics[scale=.65,page=3,trim={0cm 6.5cm 0 6.25cm},clip]{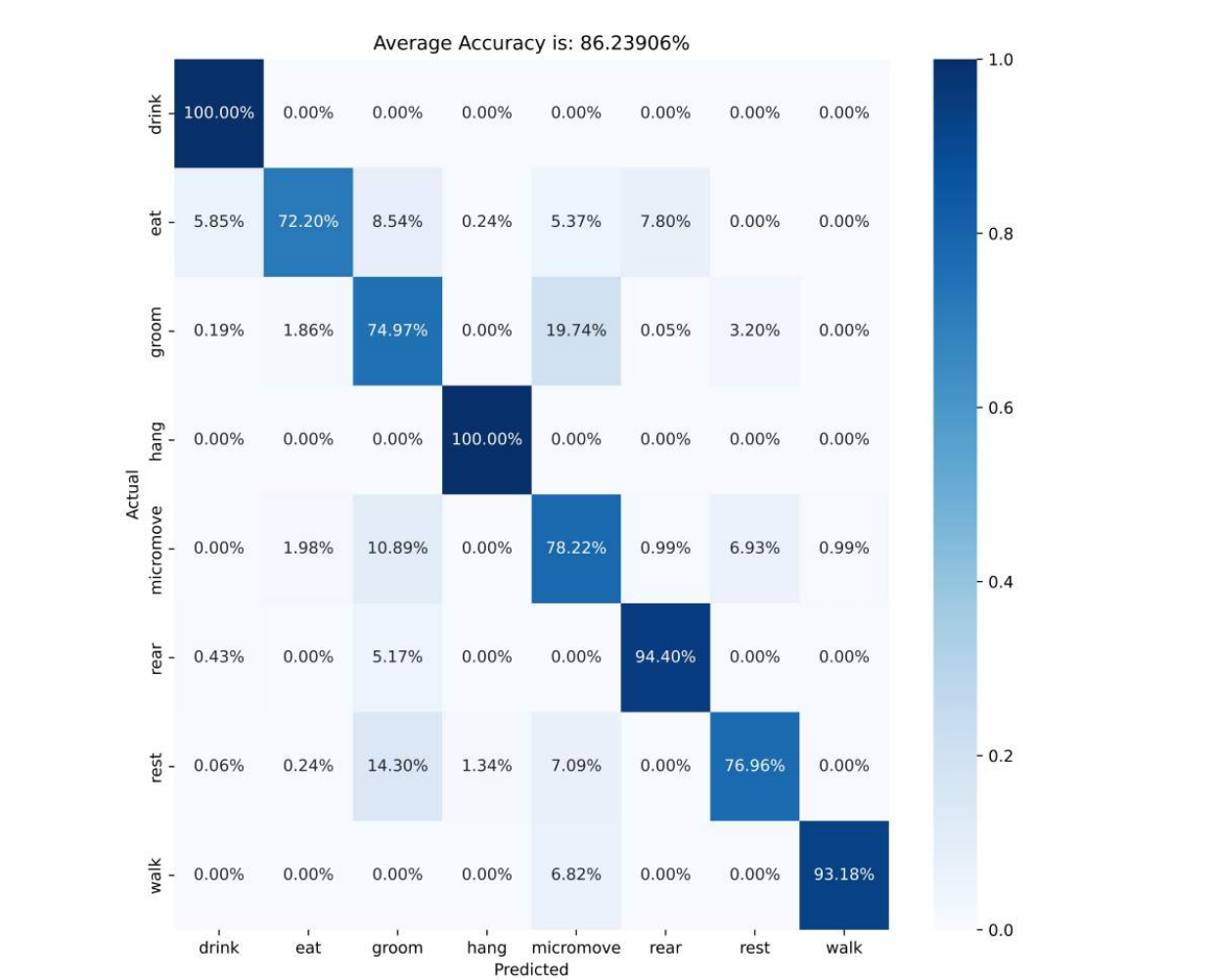}%
    \caption{Confusion Matrix for SRS + CIv3D\_BiLSTM + CIv3D\_MHA}%
    \label{cmInceptv1+CIv3D_BiLSTM+CIv3D_MHA}%
\end{figure}

\section{Original Data-Label Distributions}\label{secA3}

\begin{figure}[h]
\centering
 \subfigure[MIT mouse data]{\label{MIT}\includegraphics[width=\textwidth, page=1,]{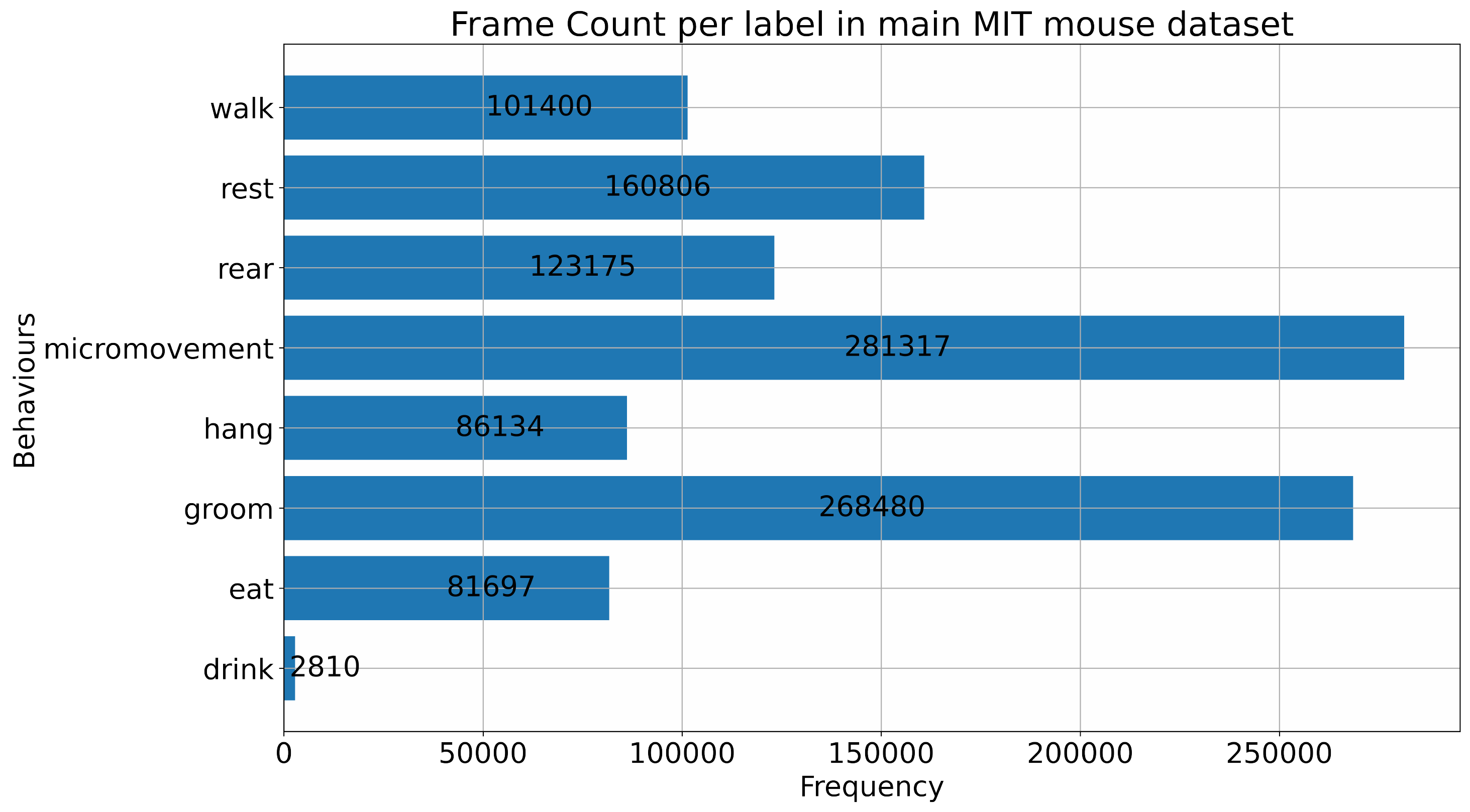}}
 
\subfigure[SCORHE mouse data]{\label{SCORHE}\includegraphics[width=\textwidth, page=2,]{figure_frameCountPlots.pdf}}
\caption{Full data summaries (before preprocesing)}
\label{fig:dataSum}
\end{figure}

\section{SCORHE AND MIT samples (after preprocessing)}\label{secA4}

\begin{figure}[h]
\centering
 \subfigure[MIT mouse data]{\label{MITRgbOftflow}\includegraphics[width=\textwidth, page=2,trim={4cm 4.5cm 3.5cm 4.5cm},clip]{figure_Images_from_mice_datasets_+_SCORHE_pred_plot.pdf}}
 
\subfigure[SCORHE mouse data]{\label{SCORHERgbOptflow}\includegraphics[width=\textwidth, page=1,trim={4cm 4.5cm 3.5cm 4.5cm},clip]{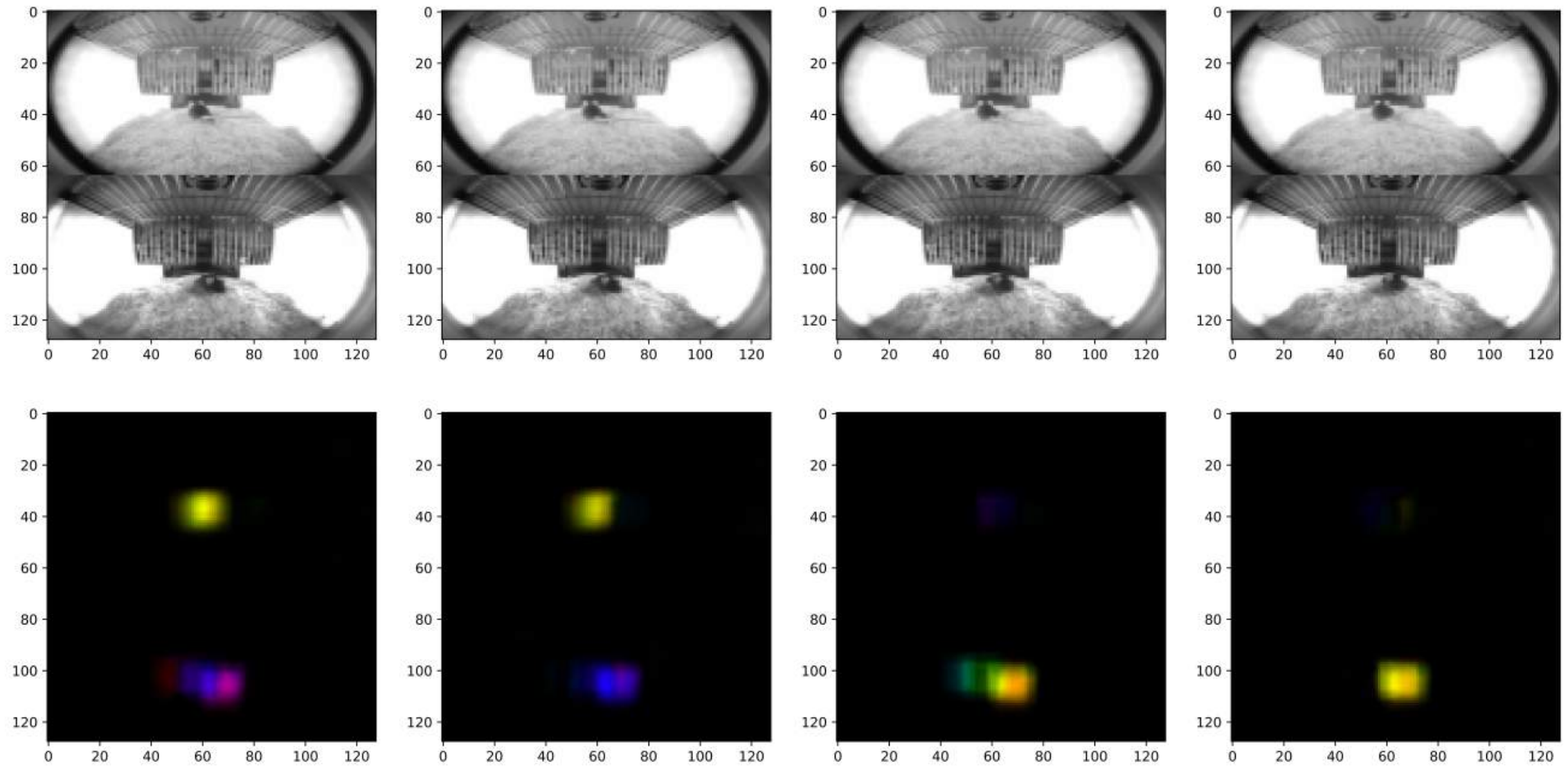}}%
\caption{Samples from SCORHE and MIT dataset}
\label{fig:dataSamples}
\end{figure}

\
\section{Schematic diagrams of SRS and CRS models}\label{secA5}
\begin{figure}[ht]%
    \centering
    \includegraphics[scale=.72,page=3,trim={2cm 2cm 0cm 2cm},clip]{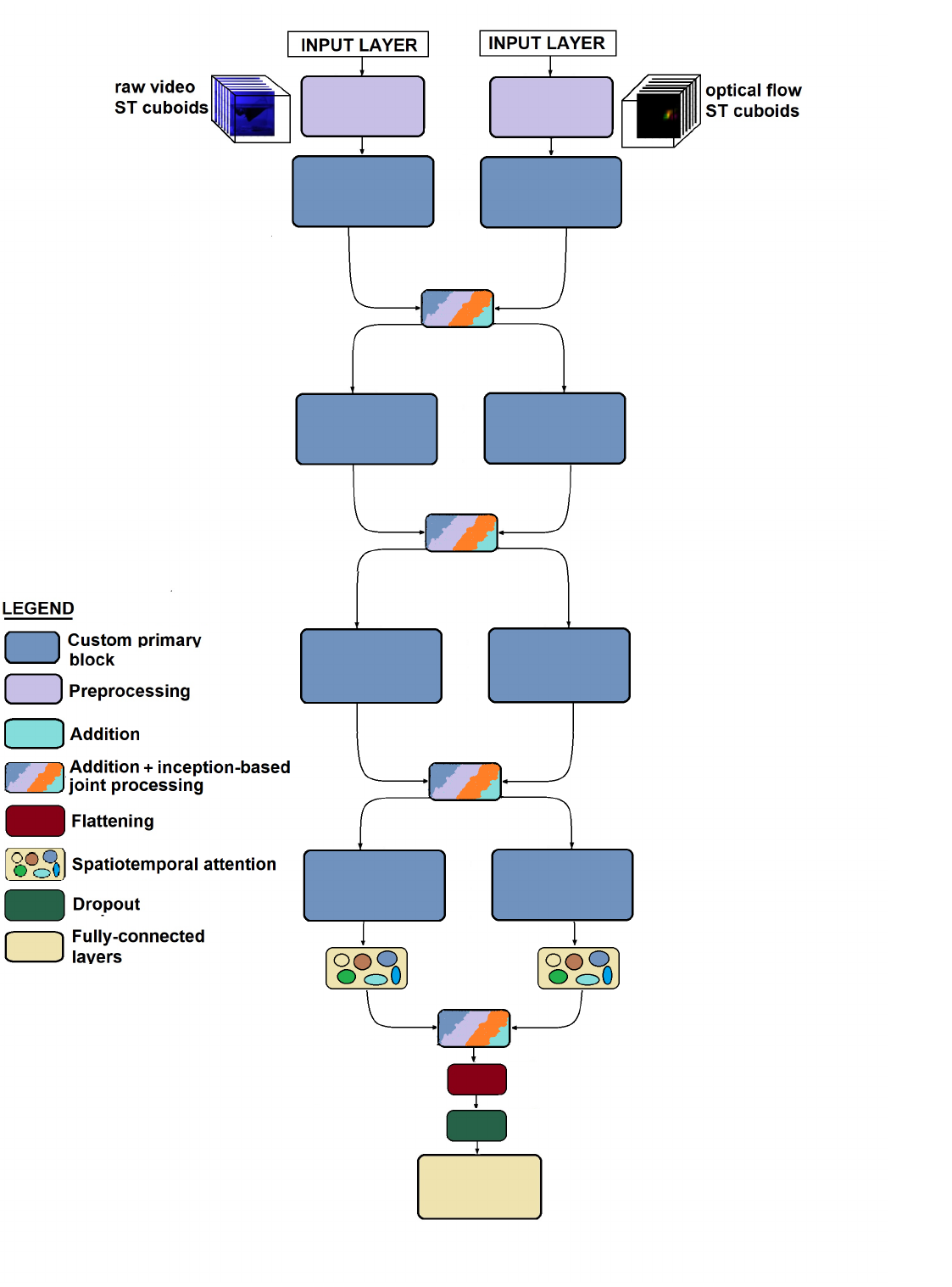}%
    \caption{Overview of CIv3D\_MHA model}%
    \label{cmhaSchematic}%
\end{figure}

The resultant feature shapes of each architecture (at major rungs) and the fully-connected sizes are in table \ref{summaryFSMods}. As mentioned in the text, the preprocessing steps for all architectures are heightwise cropping and rescaling. Furthermore, note that the ST attention block in \ref{cmhaSchematic} is the same as a single-head, single encoder stack in \citep{vaswani2017attention} but without any positional embeddings.

\begin{figure}[ht]%
    \centering
    \includegraphics[scale=.565,page=1,trim={0cm 0cm 0cm .5cm},]{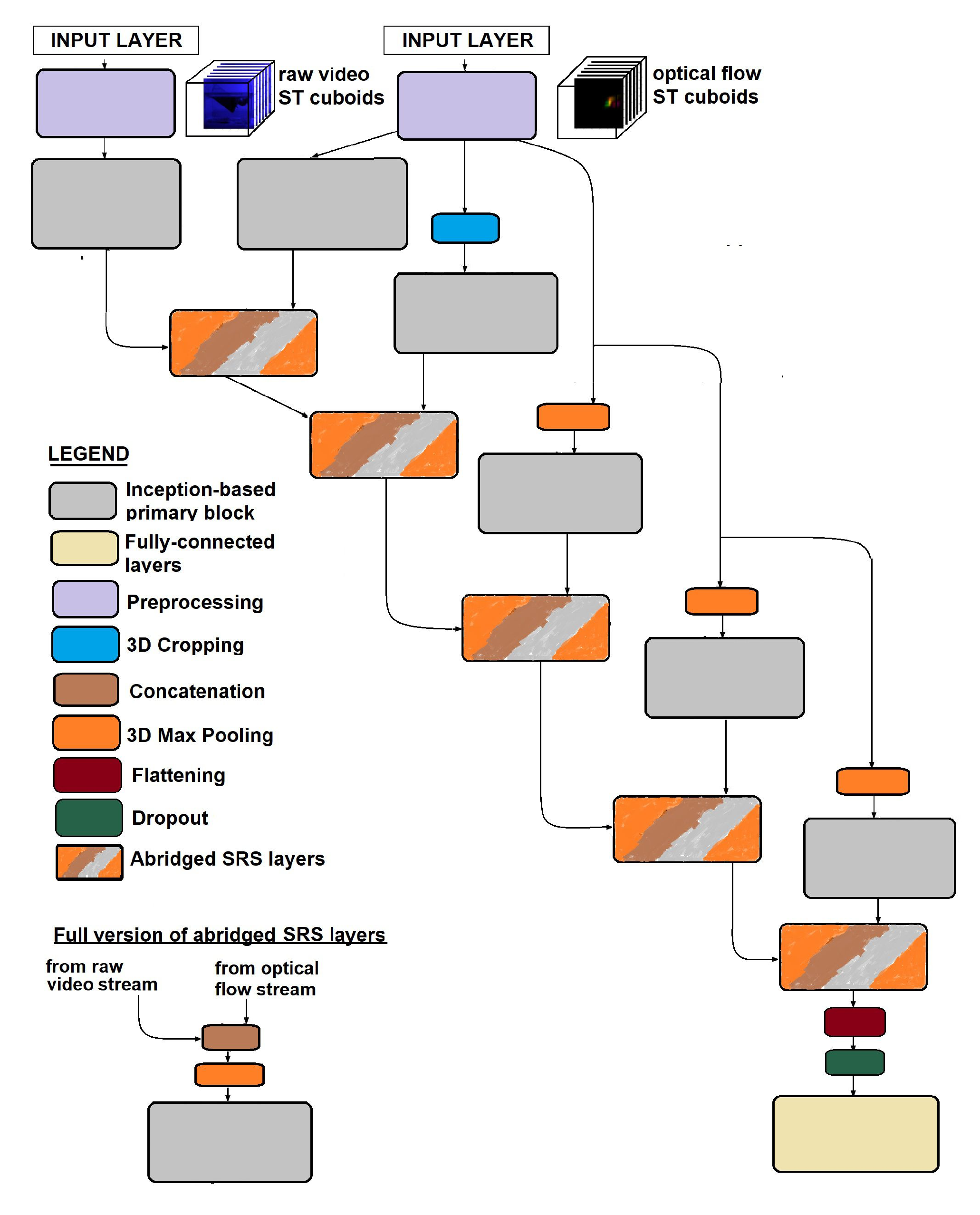}%
    \caption{Overview of SRS model}%
    \label{srsSchematic}%
\end{figure}%

\begin{figure}[ht]%
    \centering
    \includegraphics[scale=.625,page=2,trim={.5cm 0cm 0cm 0cm},]{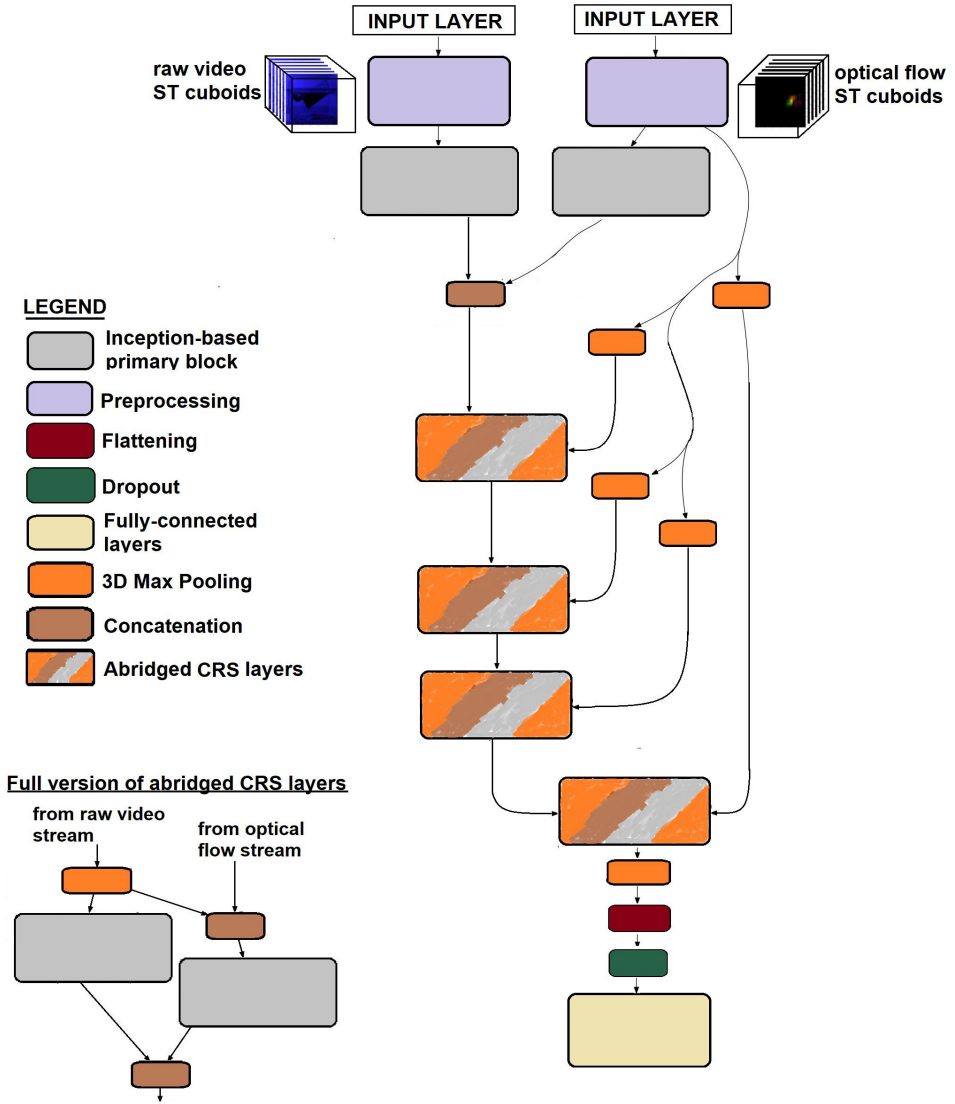}%
    \vspace*{-0.75cm} 
    \caption{Overview of CRS model}%
    \label{crsSchematic}%
\end{figure}%

\begin{figure}[ht]%
    \centering
    \includegraphics[scale=.72,page=1,trim={2cm 2cm 0cm 2cm},clip]{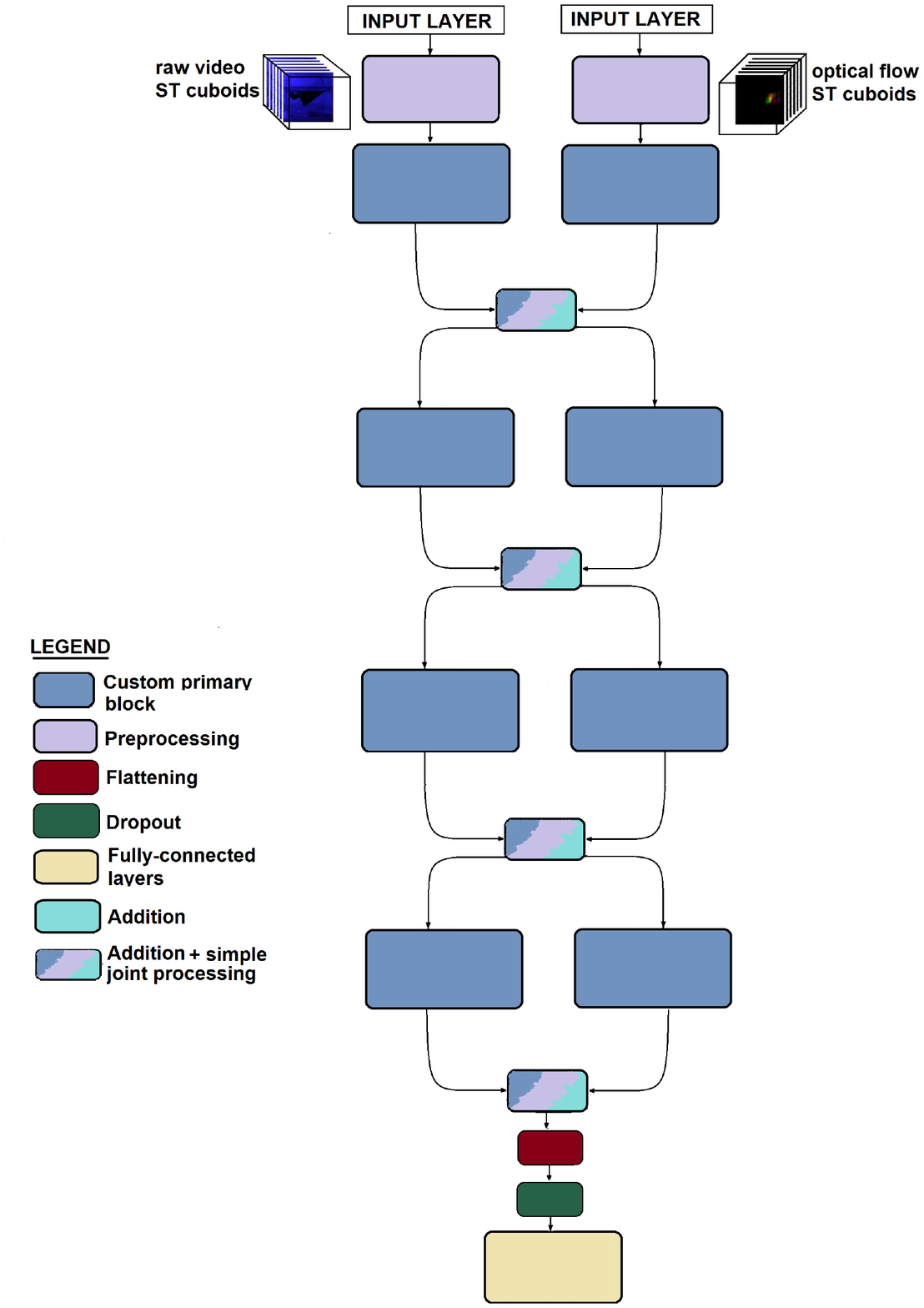}%
    \caption{Overview of Baseline model}%
    \label{baselineSchematic}%
\end{figure}

\begin{figure}[ht]%
    \centering
    \includegraphics[scale=.72,page=2,trim={2cm 2cm 0cm 2cm},clip]{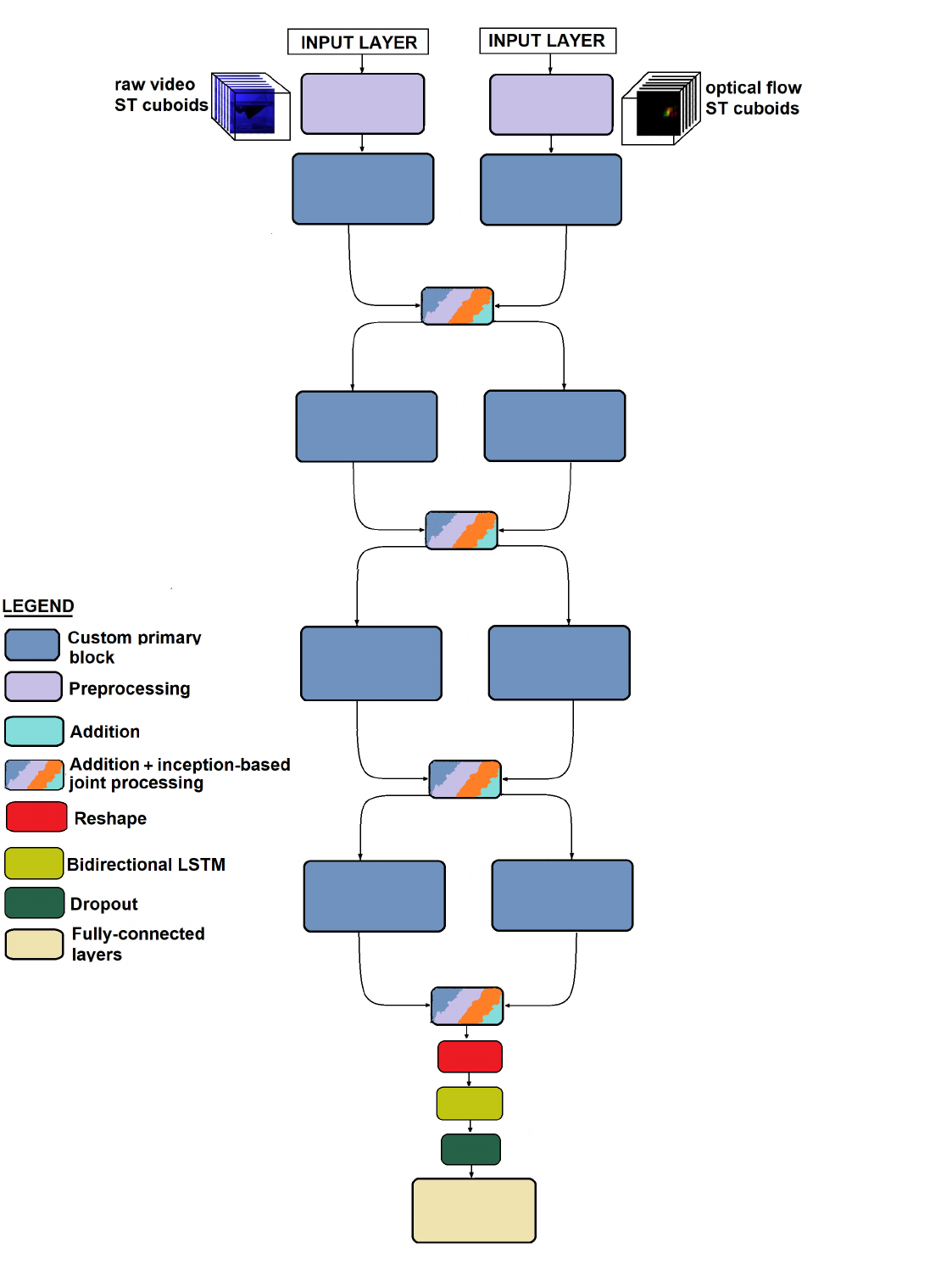}%
    \caption{Overview of CIv3D\_BiLSTM model}%
    \label{cbilstmSchematic}%
\end{figure}

\footnotetext[1]{The 'Inception-based primary block' in figures \ref{srsSchematic},\ref{crsSchematic} correspond to the \ref{pmod2}}
\footnotetext[2]{The 'Custom primary block' in figures \ref{baselineSchematic},\ref{cbilstmSchematic},\ref{cmhaSchematic} correspond to subfigure \ref{pmod1}}
\footnotetext[3]{The 'Inception-based joint processing' in figures \ref{cbilstmSchematic},\ref{cmhaSchematic} correspond to subfigure \ref{inceptv3D}}


\end{appendices}

\end{document}